\def\arxivversion{1}
\def\arxivauthornames{Zhejun Jiang \qquad Mijung Park}
\def\arxivaffiliation{The University of British Columbia}
\def\arxivpdfauthors{Zhejun Jiang; Mijung Park}
\def\arxivauthoremails{
\href{mailto:jzj03@cs.ubc.ca}{\texttt{jzj03@cs.ubc.ca}}\qquad
\href{mailto:mijungp@cs.ubc.ca}{\texttt{mijungp@cs.ubc.ca}}
}
\def\arxivacknowledgments{
\section*{Acknowledgments}
We thank the Digital Research Alliance of Canada (Compute Canada) for its
computational resources and services. Z.~Jiang and M.~Park were supported in part
by the Natural Sciences and Engineering Research Council of Canada (NSERC) and
the Canada CIFAR AI Chairs program.
}
\documentclass{article}
\usepackage{iclr2027_conference,times}

\usepackage{amsmath}
\usepackage{amssymb}
\usepackage{booktabs}
\usepackage{graphicx}
\usepackage{url}
\usepackage{xcolor}
\usepackage{enumitem}
\usepackage{array}
\usepackage{algorithm}
\usepackage{algorithmic}
\usepackage{hyperref}
\hypersetup{hidelinks}

\newcommand{\method}{\textsc{BayesER}}
\newcommand{\baseentropy}{Base Entropy}

\newcommand{\mirank}{\textsc{MIRank}}
\newcommand{\mdlm}{MDLM}

\newcommand{\maskset}{\mathcal{M}}

\usepackage{amsmath,amsfonts,bm}
\usepackage{amssymb}
\usepackage{bbm}

\def\Secref#1{Section~\ref{#1}}
\def\eqref#1{equation~\ref{#1}}
\def\Algref#1{Algorithm~\ref{#1}}

\usepackage{amsthm} %
\theoremstyle{remark}

\def\1{\bm{1}}

\newcommand{\Dat}{\mathcal{D}}

\def\vone{{\bm{1}}}

\def\vtheta{{\bm{\theta}}}
\def\vphi{{\bm{\phi}}}

\def\vv{{\bm{v}}}

\def\vx{{\bm{x}}}
\def\vy{{\bm{y}}}

\def\mA{{\bm{A}}}
\def\mB{{\bm{B}}}

\def\mF{{\bm{F}}}

\def\mI{{\bm{I}}}

\def\mW{{\bm{W}}}

\DeclareMathAlphabet{\mathsfit}{\encodingdefault}{\sfdefault}{m}{sl}
\SetMathAlphabet{\mathsfit}{bold}{\encodingdefault}{\sfdefault}{bx}{n}

\newcommand{\E}{\mathbb{E}}

\providecommand{\mathbbm}[1]{\mathbb{#1}}

\title{Bayesian Entropy-based Reordering \\
for Calibrated Diffusion Language Models}

\ifdefined\arxivversion
\author{\arxivauthornames\\
\arxivaffiliation\\
\arxivauthoremails}
\hypersetup{
  pdfauthor={\arxivpdfauthors},
  pdftitle={Bayesian Entropy-based Reordering for Calibrated Diffusion Language Models}
}

\iclrfinalcopy
\else
\author{Anonymous authors\\
Paper under double-blind review}
\fi

\begin{document}

\maketitle
\ifdefined\arxivversion
\fancyhead{}
\renewcommand{\headrulewidth}{0pt}
\fi

\begin{abstract}
Masked Diffusion Language Models (MDLMs) generate sequences by iteratively replacing masked tokens with model predictions. At each denoising step, the decoder chooses which positions are sufficiently confident to commit. Existing decoding methods typically rely on softmax confidence, which can be miscalibrated. We introduce \textit{\method{}} (\textit{BAYES}ian \textit{E}ntropy-based \textit{R}eordering), a post-hoc Bayesian decoding framework that uses predictive uncertainty to guide token commitment. In \method{}, we construct a lightweight approximate posterior centered at the pretrained checkpoint, similar to \textit{Laplace-LoRA} but without training LoRA adapters. We average predictions over posterior samples and use predictive entropy to prioritize reliable positions. We examine how posterior predictions affect position ordering and token selection across benchmarks spanning code generation, mathematical reasoning, planning, and molecular generation. We show that \method{} reduces sequence-level calibration error while preserving or improving accuracy relative to common decoding schemes, including confidence-threshold decoding.
Additionally, a posterior fitted on one code-generation dataset reduces calibration error on another without refitting, suggesting that Bayesian uncertainty may provide a transferable signal for more reliable MDLM decoding.
\end{abstract}

\section{Introduction}\label{sec:introduction}
Language models can make wrong predictions with high confidence. 
This failure is relatively prevalent, as users often mistake stylistic confidence for evidence without justified certainty. 
A confidence score should mean what it claims: predictions with confidence $q$ should be correct about a fraction $q$ of the time; this is known as \textit{calibration}. Formally, if the model $\mathbb{M}$ is perfectly calibrated, the following holds: $\mathbb{P}(\mathbb{M}'\mbox{s prediction is correct} \mid \mathbb{M}'\mbox{s confidence is } q) = q$, according to \citet{pmlr-v70-guo17a}.
This definition of calibration does not eliminate hallucinations or reasoning errors common in modern language models. However, it makes uncertainty actionable, enabling a user or downstream system to decide whether to trust an answer, verify it, defer to a human, or abstain. 

Calibration in masked diffusion language models (MDLMs)  plays an unusually direct role. Unlike autoregressive models that generate tokens from left to right, \mdlm{}s iteratively denoise a partially masked sequence. At each step, a bidirectional forward pass predicts a distribution over the vocabulary at every masked position, and the decoder selects a subset of these predictions to commit before continuing. This process enables parallel generation, arbitrary-order decoding, and infilling, and has recently scaled to models such as LLaDA and Dream \citep{nie2025llada,ye2025dream}. Consequently, confidence serves two distinct purposes in \mdlm{}s: it should indicate whether the final sequence is likely correct (the usual confidence in the model's prediction), and more interestingly, it can guide which token predictions are reliable enough to commit at a given time-step during generation.
This creates a central decoding problem that does not arise in standard autoregressive generation:
\emph{Which masked positions should be committed at each denoising step?}

Committing many positions in parallel reduces model evaluations, but once committed, those tokens provide context for all subsequent tokens being unmasked at later denoising steps. Committing too conservatively avoids such problems but reduces the computational advantage of parallel decoding. The token commitment ordering therefore controls both generation quality and inference cost. Existing decoders commonly commit to the tokens at certain positions with the largest top-token probabilities or all positions whose confidence exceeds a fixed threshold \citep{chang2022maskgit,wu2025fastdllm}. These rules are efficient and often effective, but they rely on the softmax distribution of a single parameter estimate. Neural networks are often overconfident \citep{pmlr-v70-guo17a}, and point-estimate confidence does not reveal whether a prediction remains stable under  model uncertainty.

In this paper, we investigate whether Bayesian predictive uncertainty provides a better signal for commitment ordering. Given an approximate posterior $q(\vtheta|\Dat)$ over model parameters $\vtheta$, the posterior predictive distribution over the output $\vy$ of a model averages token probabilities across posterior samples,
given a noisy state $\vx_t$ of the denoising process:
$\bar p(\vy|\vx_t)
=
\E_{\vtheta\sim q(\vtheta|\Dat)}
\left[p(\vy|\vx_t,\vtheta)\right].$ 
Its entropy, $H(\bar p(\vy|\vx_t))$, measures the total uncertainty of the Bayesian prediction. Our method, \method{} (\textbf{BAYES}ian \textbf{E}ntropy-based \textbf{R}eordering), is a post-hoc decoding framework for \mdlm{}s that uses Bayesian predictive entropy. At each denoising step, \method{} forms a posterior predictive distribution for every currently masked position, prioritizes positions with low posterior predictive entropy, selects tokens from the posterior predictive mean, and uses posterior confidence to determine how many positions to commit. In this way, \method{} adjusts both the order and pace of decoding based on the Bayesian predictive distribution.

For typical \mdlm{}s such as LLaDa \citep{nie2025llada} and Dream \citep{ye2025dream}, with 7 billion parameters or more, identifying posterior distributions over all parameters is computationally challenging. 
\textit{Laplace-LoRA} first fine-tunes LoRA adapters to obtain a task-specific maximum a posteriori solution and then approximates a posterior around the adapted model \citep{yang2024laplacelora}. 
While appropriate for downstream adaptation, this procedure changes the mean predictor and makes it difficult to distinguish the benefit of Bayesian uncertainty from that of fine-tuning. 
We instead construct a curvature-only Laplace-LoRA posterior directly \textit{at the released checkpoint}, \textit{without adapter training}. The LoRA update is zero at the posterior mean, so the deterministic model is unchanged by construction. We fit the posterior covariance with training data, representing each block of a Hessian matrix using a Kronecker
product of input and output-gradient second moments to reduce the cost of storing and expressing curvature.

Across several benchmark datasets spanning code generation, mathematical reasoning, planning, and molecule generation, \method{} substantially reduces expected calibration error (ECE) while preserving—and in several settings improving—task accuracy relative to commonly used confidence-based decoding schemes. Our results show that prior precision and the number of Laplace-LoRA layers govern a three-way trade-off among calibration, task performance, and decoding cost. We further find that a posterior constructed on one code-generation dataset can reduce calibration error on another without refitting, hinting that the learned uncertainty signal may transfer across related tasks.

\section{Background}\label{sec:background}
We first describe background information on a common decoding scheme in \mdlm{}s, Laplace-LoRA, and expected calibration error (ECE) for evaluating calibration performance. 

\subsection{Masked diffusion decoding}\label{sec:setup}

\mdlm{}s generate text by repeatedly predicting and
revealing masked tokens \citep{sahoo2024mdlm,nie2025llada,ye2025dream}. 
We denote a partially decoded, noisy sequence of length $L$ by $\vx_t^{(1:L)}$, where $\vx_t^{(l)}$ is a one-hot vector of length $|\mathcal{V}|$ (vocabulary size), representing the $ l$-th position of the sequence with one entry equaling 1 and the rest equaling zero.
We also denote which position has masked tokens at time $t$ by $\maskset_t=\{l:\vx_{t}^{(l)}=[\mathrm{MASK}]\}$. We denote a pre-trained model by $f_{\vtheta_0}$ with fixed trained parameter values $\vtheta_0$. 

In \mdlm{}s, at each time step $t$, the trained model $f_{\vtheta_0}$ outputs the token probability over the vocabulary for each position in the sequence: $p_{\vtheta_0}^{(l)}(\vv|\vx_t^{(1:L)})$. The base token is selected by choosing the argmax of this probability: $\hat y_{t}^{(l)}=\arg\max_\vv p_{\vtheta_0}^{(l)}(\vv|\vx_t^{(1:L)})$ and 
confidence is: $c_{t}^{(l)}= p_{\vtheta_0}^{(l)}(\hat y_{t}^{(l)}|\vx_t^{(1:L)})$.

\paragraph{Confidence-based decoding.}
\mdlm{}s commonly decode tokens via confidence-based decoding. 
Given a threshold $\tau$, at each denoising step $t$, $K_t$ most confident base tokens are selected by:
$K_t=\max\!\left(1,\sum_{l\in\maskset_t}\mathbbm{1}_{\{c_{t}^{(l)}\ge\tau\}}\right),$
where $\mathbbm{1}$ is an indicator function which returns $1$, if $\{c_{t}^{(l)}\ge\tau\}$, and returns $0$, if not. 
The minimum of one ensures that at least one token is unmasked at a time. 

\subsection{Laplace-LoRA}
\label{sec:background_bayes}

LoRA (Low-Rank Adaptation) parameterizes the update of each of the $Q,K,V$ matrices of a transformer architecture at each layer through two low-rank matrices
\citep{hu2021lora} (omitting the specification of $Q,K,V$ or layer index for simplicity):
$
\mW=\mW^0+\frac{\alpha}{r} \mB \mA$, with  $\mA\in\mathbb R^{r\times d_{\rm in}}$ and  $\mB\in\mathbb R^{d_{\rm out}\times r},$
where $r$ is a rank of the low-rank matrices and $\alpha$ is a scaling hyper-parameter. 

\paragraph{Laplace-LoRA} 
To gain posterior uncertainty about the model, Laplace-LoRA first optimizes the LoRA adapter parameters \(\mA\) and \(\mB\) on downstream tasks, then places a Gaussian distribution over \(\mA\) and \(\mB\), while keeping the underlying weights $\mW^0$ fixed \citep{yang2024laplacelora}. 
Laplace-LoRA uses the \textit{Laplace} approximation to estimate the covariance of the adapter parameters by learning 
the curvature of the negative log posterior at a posterior mode
\citep{mackay1992laplace}. 
For decent-sized large language models, computing the exact curvature is intractable. 
Laplace-LoRA proposes two levels of approximations: (a) rather than computing the exact curvature, it uses the Fisher information, $\mF:= \sum_{n=1}^{N}
\mathbb{E}_{y \sim P\left(\cdot \mid f_{\boldsymbol{\theta}}(\mathbf{x}_n)\right)}
\left[
\nabla_{\boldsymbol{\theta}}
\log P\left(y \mid f_{\boldsymbol{\theta}}(\mathbf{x}_n)\right)
\left(
\nabla_{\boldsymbol{\theta}}
\log P\left(y \mid f_{\boldsymbol{\theta}}(\mathbf{x}_n)\right)
\right)^{\top}
\right]$ ; (b) furthermore,  the Fisher information is computed on each linear layer separately using \textit{Kronecker-factored approximate curvature (KFAC)}, forming a block structure of the Fisher information matrix. 
KFAC represents each layer's Fisher information using a Kronecker product of input covariance and output covariance,
reducing the cost of storing and manipulating curvature
\citep{martens2015kfac,ritter2018scalable}.
Our method, \method{}, also specifies the curvature using these approximations, but without fine-tuning adapters, but at a fixed checkpoint as shown in \Secref{sec:method-posterior}.

\subsection{Calibration and sequence-level confidence}
\label{sec:background_calibration}

Expected calibration error (ECE) \citep{naeini2015obtaining} measures how well the model's prediction accuracy matches its confidence. Large ECE indicates a mismatch between confidence and accuracy, reflecting overconfidence or underconfidence.
With $N$ number of samples, predicted labels $\hat{y}_n$ for each sample $n$, and true label $y_n$, ECE groups confidences into $M$ bins (each bin is $\mathcal B_m$) and measures
\begin{align}
\label{eq:ece}
ECE &= \sum_{m=1}^M \frac{|\mathcal B_m|}{N} \mid acc( \mathcal B_m) - conf(\mathcal B_m)\mid, 
\end{align} where $acc(\mathcal B_m) = \frac{1}{|\mathcal B_m|}\sum_{n \in \mathcal B_m} \vone(\hat{y}_n =y_n),  conf(\mathcal B_m)= \frac{1}{|\mathcal B_m|}\sum_{n \in \mathcal B_m} \hat{p}_n$, and $\hat{p}_n$ is the model's confidence for sample $n$: $\hat{p}_n = \max_c p(y_n=c|x_n)$. Directly applying ECE to text \textit{sequences} is not straightforward. In \Secref{sec:method-sequence-confidence}, we show two ways to evaluate ECE on the generated sequence.

Another commonly used probabilistic prediction metric which we will also use is the \textit{Brier score} \citep{brier1950verification}. For our binary task-correctness evaluation,
$\mathrm{BS} = \frac{1}{N} \sum_{n=1}^N (c_n-z_n)^2$, where $c_n$ is the sequence confidence score and $z_n\in\{0,1\}$ indicates task correctness.

\section{Method: \method{}}\label{sec:method}

A decoder in \mdlm{}s makes three decisions at every denoising step:
\emph{which positions to commit}, \emph{which tokens to write}, and
\emph{how many positions to commit}.
\method{} makes all three decisions from a checkpoint-centered posterior predictive distribution.
Specifically, it ranks positions by posterior predictive entropy, selects token identities from the posterior predictive mean, and determines the commitment budget from posterior confidence.

\method{} consists of two stages: (1) First, we fit the posterior distribution over LoRA adapter parameters using training data, from which we obtain the posterior samples; and (2) Then, using the posterior samples, we compute the predictive posterior entropy at every denoising step $t$ to determine how many tokens to unmask and which positions to fill with which token. We describe each of these two stages below. 

\subsection{Phase I: Curvature-Only Laplace-LoRA Posterior}
\label{sec:method-posterior}

\paragraph{Checkpoint-centered LoRA parameterization.}
We place LoRA adapters on selected query and value matrices in the selected \(J\) transformer layers.
Below, we omit the indices for $Q$ and $V$ and show only the layer index $j$, since the LoRA operation is the same for $Q$ and $V$. 

For a selected pretrained matrix \(\mW_{j,0}\in\mathbb{R}^{d_{\mathrm{out}}\times d_{\mathrm{in}}}\), the effective weight is
\begin{equation}
    \mW_j(\phi_j)
    =
    \mW_{j,0}
    +
    \frac{\alpha}{r} \mB_j \mA_j,
    \qquad
    \mA_j\in\mathbb{R}^{r\times d_{\mathrm{in}}},
    \quad
    \mB_j\in\mathbb{R}^{d_{\mathrm{out}}\times r},
    \label{eq:bayeser-lora}
\end{equation}
where 
we let 
    $\vphi_0
    =
    \{\mA_{j,0},\mB_{j,0}\}_{j},$
    and $\mB_{j,0}=0,$
with \(\mA_{j,0}\) given by the standard LoRA initialization.
Because \(\mB_{j,0}\mA_{j,0}=0\), the posterior-mean model is exactly the released checkpoint:
\[
    f_{\vtheta_0,\vphi_0}=f_{\vtheta_0}.
\]
Unlike standard Laplace-LoRA, \textit{\textbf{we do not optimize the adapter means on a downstream objective.}}
We use the training data only to estimate local curvature.

\paragraph{Approximating Fisher information with KFAC}
For each LoRA adapter matrix index $m$, where $m$ can vary between layers and $Q$ and $V$, we denote its input activation for each observation $\widetilde{\mathbf{x}}_n$ by \(h_{m,n}\), and let
\[
    g_{m,n}
    =
    \nabla_{a_{m,n}}
    \log p_{\vtheta_0,\vphi_0}
    \bigl(Y_n\mid \widetilde{\mathbf{x}}_n\bigr)
\]
be the gradient with respect to the corresponding pre-activation \(a_{m,n}\), where
\(Y_n\) is distributed according to the model predictive distribution.
We approximate the Fisher information matrix separately for each adapter matrix, using Kronecker-factored approximate curvature (KFAC):
\begin{align}
    F_m \approx
    N_m\left(G_m\otimes C_m\right), \quad \mbox{ where }
    C_m
    =
    \mathbb{E}\!\left[h_mh_m^\top\right], \mbox{ and }
    \;
    G_m
    =
    \mathbb{E}\!\left[g_mg_m^\top\right],
\end{align}
where \(N_m\) counts the contributing source-position observations represented by the sampled rows for block \(m\), rather than calibration sequences or masked targets.
The factor \(C_m\) captures correlations among matrix inputs, whereas \(G_m\) captures correlations among output-gradient coordinates.

Using an isotropic Gaussian prior precision \(\lambda\), we construct the checkpoint-centered, Laplace-style Gaussian approximation
\begin{equation}
    q_\lambda(\vphi)
    =
    \mathcal{N}
    \left(
        \vphi_0,\,
        \left[
            \operatorname{blockdiag}(F_1,\ldots,F_R)+\lambda I
        \right]^{-1}
    \right),
    \label{eq:bayeser-posterior}
\end{equation} where $R$ is the total number of individual LoRA adapter matrices across the $J$ we consider. For instance, if we have $Q, V$ matrices per layer and apply LoRA to 10 layers, the posterior covariance consists of small, manageable $40$ blocks (since each $Q$ and $V$ has two adapter matrices, $\mA$ and $\mB$). 
Together, these blocks define a single joint Gaussian over all selected adapter parameters. Blocks are independent across matrices, projections, and layers; however, within-matrix KFAC correlations remain, and the leading eigendirections of large factors are retained. Directions omitted by the truncated eigenspace retain prior variance.

\paragraph{Uncertainty in the LoRA factors}
At the initialization $\vphi_0$, we have $\mB_{j, 0}$ initialized as 0. Changing $\mA_{j, 0}$ alone therefore does not affect $\mW_j(\vphi_j)$, so the gradient with respect to $\mA_j$ and its Fisher block is zero. Under our Gaussian approximation, this gives:
$   \operatorname{vec}(\mA_j)
    \sim
    \mathcal{N}\!\left(
        \operatorname{vec}(\mA_{j,0}),\,\lambda^{-1}\mI
    \right).$
Therefore, uncertainty in $\mA_j$ is entirely prior-controlled, while uncertainty in $\mB_j$ depends on both the calibration curvature and the prior.
At prediction time, each posterior draw $\vphi^{(s)} \sim q_\lambda(\vphi)$ gives a
corresponding pair of sampled LoRA matrices $\mA_j^{(s)}$ and $\mB_j^{(s)}$. We use these matrices to construct the sampled model weights: 
$
    \mW_j(\vphi_j^{(s)})
    =
    \mW_{j,0}
    + \frac{\alpha}{r}\mB_j^{(s)}\mA_j^{(s)}$.
Although $\mA_j$ has zero local Fisher curvature at initialization, its sampled variation still affects predictions because $\mB_j^{(s)}$ is generally non-zero. Writing $\mA_j^{(s)}=\mA_{j,0}+\delta \mA_j^{(s)}$ and $\mB_j^{(s)}=\delta \mB_j^{(s)}$, the sampled update includes the bilinear term $\frac{\alpha}{r}\delta \mB_j^{(s)}\delta \mA_j^{(s)}$ in addition to $\frac{\alpha}{r}\delta \mB_j^{(s)}\mA_{j,0}$. Thus, predictive variation reflects both data-informed curvature and prior-controlled variation in locally unconstrained directions.

\begin{algorithm}[t]
\caption{\method{}}
\label{alg:bayeser}
\begin{algorithmic}[1]
\REQUIRE Pre-trained model:$f_{\vtheta_0}$, confidence threshold: $\tau_{\mathrm{B}}\in[0,1]$,
how many posterior samples to draw:$S\ge1$,
initial sequence $\mathbf{x}_T^{(1:L)}$,
and the number of selected tokens per position:$K_v\geq1$
\STATE \textbf{Phase I:} Fit the parameter posterior $q_\lambda$ using \Algref{alg:lora_fit}
\STATE \qquad \quad\; Then, sample parameters $\vtheta^{(s)}\sim q_\lambda$ for $s=1,\ldots,S$
\STATE \textbf{Phase II:}
\STATE $t\leftarrow T$; $\mathcal C\leftarrow\emptyset$; $\maskset_T\leftarrow\{l:\vx_{T}^{(l)}=[\mathrm{MASK}]\}$
\WHILE{$\maskset_t\ne\emptyset$}
    \STATE Run base model $f_{\vtheta_0}(\mathbf{x}_t^{(1:L)})$ to obtain token logits and confidences $c_{t}^{(l)}$ for each position $l$
    \STATE For each masked position $l$, define a set $\mathcal V_{t}^{(l)}$ containing the $K_v$ highest-scoring tokens
    \FOR{$s=1,\ldots,S$}
        \STATE Given each posterior sample, run the sampled model $f_{\vtheta^{(s)}}(\vx_t^{(1:L)})$
        \STATE $p_{t,s}^{(l)}\leftarrow$ softmax of sampled-model logits over tokens in $\mathcal V_{t}^{(l)}$ 
    \ENDFOR
    \STATE $\bar p_{t}^{(l)}\leftarrow S^{-1}\sum_s p_{t,s}^{(l)}$
    \STATE $H_{t}^{(l)}\leftarrow-\sum_{v\in\mathcal V_{t}^{(l)}}\bar p_{t}^{(l)}(v)\log\bar p_{t}^{(l)}(v)$
    \STATE $\hat y_{t}^{(l)}\leftarrow\arg\max_{v\ne[\mathrm{MASK}]}\bar p_{t}^{(l)}(v)$; and 
    $c_{t}^{(l)}\leftarrow\bar p_t^{(l)}(\hat y_{t}^{(l)})$
    \STATE $K_t\leftarrow\max(1,\sum_{l\in\maskset_t}\mathbbm{1}_{c^{(l)}_{t}\ge\tau_{\mathrm{B}}})$
    \STATE $\mathcal P_t\leftarrow$ the $K_t$ masked positions with lowest entropy $H_{t}^{(l)}$ 
    \STATE Record $(t,l,\hat y^{(l)}_{t},c^{(l)}_{t})$ in $\mathcal C$ for each $l\in\mathcal P_t$
    \STATE $\vx_{t-1}^{(l)}\leftarrow\hat y^{(l)}_{t}$ for each $l\in\mathcal P_t$
    \STATE $t\leftarrow t-1$
    \STATE $\maskset_t\leftarrow\{l:\vx_{t}^{(l)}=[\mathrm{MASK}]\}$
\ENDWHILE
\RETURN Updated sequence $\vx_t^{(l)}$ and saved tokens and their probabilities $\mathcal C$
\end{algorithmic}
\end{algorithm}

\subsection{Phase II: Posterior Predictive Entropy for Position Ordering}
\label{sec:method-entropy}

At step \(t\), the pre-trained model produces base logits
\(\ell^{(l)}_{0,t}\in\mathbb{R}^{|\mathcal{V}|}\)
for each location $l$.
For each position, we define a candidate set containing the top-$K_v$ highest-scoring non-masked tokens, i.e., 
$\mathcal{V}_{t}^{(l)}=\operatorname{TopK}_{v\in\mathcal{V}\setminus\{\texttt{[MASK]}\}}
    \left(\ell^{(l)}_{0,t}(v);K_v\right).$
For \(S\) independent posterior draws
\(\vphi^{(s)}\sim q_\lambda(\vphi)\),
let \(\ell^{(l)}_{t,s}\) be the sampled-model logit.
We first normalize each draw over the candidate set $\mathcal{V}_{t}^{(l)}$:
$    p^{(l)}_{t,s}(v)
    =
    \frac{\exp\!\left(\ell^{(l)}_{t,s}(v)\right)}
    {\sum_{u\in\mathcal{V}_{t}^{(l)}}
     \exp\!\left(\ell^{(l)}_{t,s}(u)\right)},$ for $v\in\mathcal{V}_{t}^{(l)}$.
Importantly, we average probabilities rather than logits:
  $  \overline{p}_{t}^{(l)}(v)
    =
    \frac{1}{S}
    \sum_{s=1}^{S}
    p^{(l)}_{t,s}(v).$
Our position-ordering criterion is the posterior predictive entropy given by 
\begin{equation}
    H_{t}^{\mathrm{pred},(l)}
    =
    H\!\left(\overline{p}_{t}^{(l)}\right)
    =
    -\sum_{v\in\mathcal{V}_{t}^{(l)}}
    \overline{p}_{t}^{(l)}(v)
    \log \overline{p}_{t}^{(l)}(v).
    \label{eq:bayeser-predictive-entropy}
\end{equation}
Low entropy indicates that the posterior predictive distribution concentrates on a small number of candidates, making the position comparatively safe to commit.
High entropy indicates that no clear token is supported under the current partially masked context, so the position should be deferred.

\paragraph{Commitment Budgets}

Given a posterior-confidence threshold \(\tau_B\),  \method{} sets its per-step commitment budget to
    $K_t
    =
    \max\left(
        1,\,
        \sum_{l\in\mathcal{M}_t}
        \mathbbm{1}_{        \left\{
            c^{(l)}_{t}\ge\tau_B
        \right\}}
    \right).$
It then chooses the \(K_t\) positions with the lowest posterior predictive entropy:
\begin{equation}
    \mathcal{P}_t
    =
    \underset{
        \substack{
            \mathcal{P}\subseteq\mathcal{M}_t\\
            |\mathcal{P}|=K_t
        }
    }{\arg\min}
    \sum_{l\in\mathcal{P}}
    H^{\mathrm{pred}(l)}_{t}.
    \label{eq:bayeser-position-set}
\end{equation}
For each \(l\in\mathcal{P}_t\), we commit
$   \vx_{t-1}^{(l)}
    =
    \widehat{y}^{(l)}_{t}, \mbox{ where } \hat y_{t}^{(l)}\leftarrow\arg\max_{v\ne[\mathrm{MASK}], }\bar p_{t}^{(l)}(v)$, 
while all unselected positions remain unchanged.
\Algref{alg:bayeser} summarizes our algorithm.

\subsection{Evaluating Sequence-Level Confidence}
\label{sec:method-sequence-confidence}

To evaluate ECE on a generated sequence, we first compute the geometric mean as a length-normalized summary of commitment confidence, for a token scope \(\mathcal{S}\subseteq\mathcal{C}\): 
\begin{equation}
 \pi_{\mathrm{seq}}(\mathcal{S})
    =
    \exp\left(
        \tfrac{1}{|\mathcal{S}|}
        \sum_{(t,l)\in\mathcal{S}}
        \log
        c^{(l)}_{t}
    \right).
    \label{eq:bayeser-sequence-confidence}
\end{equation}
using the token probability available at the moment of commitment, before the committed token becomes part of the context for later denoising steps, denoted by 
$\mathcal{C}
    =
    \left\{
        \bigl(t,l,y^{(l)}_{t},
        c^{(l)}_{t}\bigr)
    \right\}$, with
$ c^{(l)}_{t}
    =
    \bar{p}_{t}^{(l)}
    \left(y^{(l)}_{t}\right).$
This aggregation is a candidate confidence estimator, which 
we use for \(\pi_{\mathrm{seq}}(\mathcal{S}_n)\) as the confidence \(\hat{p}_n\) for each generated output and compute ECE using ~\eqref{eq:ece}, with correctness determined by the benchmark's task criterion.

In \Secref{sec:experiments}, 
we report the score over two scopes: all generated-content tokens (\(\mathrm{ECE}_{\mathrm{gen}}\)) and task-specific final-answer tokens (\(\mathrm{ECE}_{\mathrm{ans}}\)).
Note that low \(\mathrm{ECE}_{\mathrm{gen}}\) does not establish that every sentence in a reasoning trace is correct. It means that confidence aggregated over that trace agrees, at the resolution measured, with the frequency of final task success.
The generated-content and final-answer variants use different token scopes but share the same task-level correctness event.

This geometric mean is different from \citet{li2026confidence}, where their ECE is an arithmetic mean of token probabilities extracted from a single final forward pass under teacher forcing on the completed sequence.
Our geometric mean summarizes the model’s confidence at the commitment decisions made along the decoding trajectory. 

\subsection{Mutual Information and Base Entropy}

\paragraph{Mutual Information}
For comparison, one might consider the token--parameter mutual information (MI) given by 
$    I_{t}^{(l)}
    =
    H^{\mathrm{pred}(l)}_{t}
    -
    \frac{1}{S}
    \sum_{s=1}^{S}
    H\!\left(p^{(l)}_{t,s}\right).$
MI isolates disagreement among posterior draws \citep{houlsby2011bald}, but it can be small even when every draw assigns a diffuse token distribution.
This makes MI insufficient for deferred commitment: posterior samples may agree that a position remains ambiguous because the current state does not yet provide enough context.
We therefore use total posterior predictive entropy as the default ordering score and retain MI-based ordering as an ablation. We empirically compare MI and predictive entropy in Appendix~\ref{app:mi_entropy_grid}. 

\paragraph{Base Entropy}
To distinguish the effect of entropy-based ordering from that of Bayesian marginalization, we can also compare our predictive entropy-based selection from base entropy given by: $H^{(l)}_{0,t}=H\!\left(p^{(l)}_{0,t}\right)$, where 
$    p^{(l)}_{0,t}(v)
    =
    \frac{\exp\!\left(\ell^{(l)}_{0,t}(v)\right)}
    {\sum_{u\in\mathcal{V}_{t}^{(l)}}
     \exp\!\left(\ell^{(l)}_{0,t}(u)\right)}.$
This baseline, \textsc{Base Entropy}, ranks positions by \(H^{(l)}_{0,t}\) while retaining the model's tokens and confidence-based commitment budget.

\makeatletter\setlength{\@fptop}{0pt}\makeatother

\section{Related Work}\label{sec:related}
\paragraph{Decoding and unmasking policies.}
MaskGIT and Fast-dLLM
\citep{chang2022maskgit,wu2025fastdllm} present confidence-based decoders ; Deferred Commitment Decoding adapts
the active window \citep{shu2026dcd}, and Entropy based sampler uses a point estimate of entropy to control
parallel commitments \citep{benhamu2025entropy}. Lookahead Unmasking and the
Information-Gain (Info-Gain) Sampler evaluate candidate successor states
\citep{lee2025lookahead,yang2026infogain}. Other approaches learn unmasking
policies through reinforcement learning \citep{jazbec2025learning} or derive
ordering corrections from terminal rewards, as in DPRM \citep{bu2026dprm}.
None of the above methods uses \textit{posterior predictive uncertainty} to choose positions, tokens,
and commitment counts. 

\paragraph{Bayesian inference for pretrained models.}
Laplace-LoRA fits a posterior around fine-tuned adapters
\citep{yang2024laplacelora}, BLoB \citep{wang2024blob} learns a variational posterior during adapter
training , and Training-Free Bayesianization \citep{shi2024tfb} assigns
uncertainty to already-trained adapters. Building on
structured Laplace approximations \citep{ritter2018scalable,daxberger2021laplace},
our curvature-only construction fits uncertainty around zero-update adapters
at the released checkpoint. This preserves the deterministic mean model and
allows us to study posterior prediction without conflating it with downstream fine-tuning.

\paragraph{Uncertainty in generation.}
For continuous diffusion, BayesDiff propagates uncertainty using last-layer
Laplace \citep{kou2023bayesdiff}, while \citet{jazbec2025generative} develop a
Bayesian measure for assessing synthetic samples. Our focus is utilizing the uncertainty at
unresolved discrete token positions to guide commitment during generation.

\section{Experiments}\label{sec:experiments}
\begin{table}[t]
\caption{Evaluation on mathematical reasoning and programming tasks. Accuracy is reported in percent; ECE and Brier values and their standard deviations are multiplied by 100. Subscripts are the standard deviations of three seeds on the corresponding column scale. Numbers in parentheses are relative percentage improvements over confidence-thresholding decoding. Bold values indicate the best result and underlined values indicate the second-best result per metric within each benchmark.}
\label{tab:main_results}
\centering
\scriptsize
\setlength{\tabcolsep}{4pt}
\setlength{\extrarowheight}{1pt}
\newcommand{\resultdet}[2]{\ensuremath{#1\,(#2)}}
\newcommand{\resultseed}[3]{\ensuremath{#1_{\scriptscriptstyle #2}\,(#3)}}
\resizebox{\textwidth}{!}{%
\begin{tabular}{@{}llrrrr@{}}
\toprule
Dataset & Method & Accuracy (\%) $\uparrow$ & ECE$_{\mathrm{gen}}$ ($\times100$) $\downarrow$ & ECE$_{\mathrm{ans}}$ ($\times100$) $\downarrow$ & Brier$_{\mathrm{gen}}$ ($\times100$) $\downarrow$ \\
\midrule
\textbf{MBPP} & Threshold (Fast-dllm) & $\mathbf{57.60}$ & 29.18 & 31.24 & $31.85$ \\
 & \baseentropy & \resultdet{57.00}{-1.04} & \resultdet{28.99}{+0.65} & \resultdet{31.53}{-0.93} & \resultdet{31.23}{+1.97} \\
 & Info-gain & \resultseed{54.00}{1.71}{-6.25} & \resultseed{31.60}{1.71}{-8.29} & \resultseed{33.73}{1.71}{-7.97} & \resultseed{33.15}{1.23}{-4.07} \\
 & \method{} (best accuracy) & \resultseed{\underline{57.53}}{0.12}{-0.12} & \resultseed{\underline{23.54}}{0.11}{+19.33} & \resultseed{\underline{25.70}}{0.09}{+17.73} & \resultseed{\underline{28.40}}{0.02}{+10.84} \\
 & \method{} (best calibration) & \resultseed{56.73}{1.01}{-1.51} & \resultseed{\mathbf{16.32}}{1.21}{+44.07} & \resultseed{\mathbf{18.17}}{1.02}{+41.84} & \resultseed{\mathbf{25.57}}{0.36}{+19.73} \\
\midrule
\textbf{HumanEval} & Threshold (Fast-dllm) & 54.88 & 30.90 & 30.59 & $33.27$ \\
 & \baseentropy & \resultdet{\mathbf{57.32}}{+4.45} & \resultdet{27.25}{+11.81} & \resultdet{26.71}{+12.68} & \resultdet{31.80}{+4.42} \\
 & Info-gain & \resultseed{51.22}{0.61}{-6.67} & \resultseed{33.68}{0.57}{-9.00} & \resultseed{33.28}{0.53}{-8.79} & \resultseed{35.82}{0.82}{-7.66} \\
 & \method{} (best accuracy) & \resultseed{\underline{56.30}}{3.01}{+2.59} & \resultseed{\underline{24.77}}{3.04}{+19.84} & \resultseed{\underline{24.61}}{3.13}{+19.55} & \resultseed{\underline{30.26}}{1.80}{+9.05} \\
 & \method{} (best calibration) & \resultseed{55.08}{1.96}{+0.36} & \resultseed{\mathbf{20.16}}{2.04}{+34.76} & \resultseed{\mathbf{20.33}}{1.98}{+33.54} & \resultseed{\mathbf{27.76}}{1.05}{+16.56} \\
\midrule
\textbf{GSM8K} & Threshold (Fast-dllm) & 76.19 & 10.01 & 17.91 & $17.46$ \\
 & \baseentropy & \resultdet{\mathbf{78.17}}{+2.60} & \resultdet{\underline{7.14}}{+28.67} & \resultdet{\underline{15.98}}{+10.78} & \resultdet{\mathbf{16.39}}{+6.13} \\
 & Info-gain & \resultseed{69.70}{0.44}{-8.52} & \resultseed{10.17}{0.18}{-1.60} & \resultseed{18.80}{0.06}{-4.97} & \resultseed{16.62}{0.13}{+4.81} \\
 & \method{} (best accuracy) & \resultseed{\underline{77.98}}{0.38}{+2.35} & \resultseed{7.17}{0.41}{+28.37} & \resultseed{16.10}{0.20}{+10.11} & \resultseed{\underline{16.53}}{0.18}{+5.31} \\
 & \method{} (best calibration) & \resultseed{76.89}{0.52}{+0.92} & \resultseed{\mathbf{6.50}}{0.55}{+35.06} & \resultseed{\mathbf{15.75}}{0.52}{+12.06} & \resultseed{16.98}{0.26}{+2.73} \\
\midrule
\textbf{MATH500} & Threshold (Fast-dllm) & $\underline{42.40}$ & $\underline{45.50}$ & $\underline{50.08}$ & $\underline{45.33}$ \\
 & \baseentropy & \resultdet{40.00}{-5.66} & \resultdet{46.27}{-1.69} & \resultdet{53.31}{-6.45} & \resultdet{45.77}{-0.98} \\
 & Info-gain & \resultseed{40.73}{0.58}{-3.94} & \resultseed{45.96}{0.70}{-1.01} & \resultseed{52.04}{1.13}{-3.91} & \resultseed{45.72}{0.46}{-0.87} \\
 & \method{} (best accuracy) & \resultseed{\mathbf{43.53}}{1.17}{+2.67} & \resultseed{\mathbf{42.71}}{1.12}{+6.13} & \resultseed{\mathbf{49.55}}{1.09}{+1.06} & \resultseed{\mathbf{43.19}}{0.84}{+4.72} \\
\bottomrule
\end{tabular}%
}
\end{table}

\subsection{Setup}

We evaluate Dream-v0-Instruct-7B \citep{ye2025dream} on code generation (MBPP, HumanEval), mathematical reasoning (GSM8K, MATH500), and planning (4$\times$4 Sudoku, Countdown). HumanEval transfers the MBPP posterior without refitting. 
We also include LLaDA results in Appendix~\ref{app:llada10}. 
We additionally evaluate molecular scaffold decoration with GenMol V2. 
Detailed configurations are available in Appendix~\ref{app:protocol_details}.

We compare against confidence-threshold decoding, \baseentropy{}, and Info-gain \citep{yang2026infogain}. Info-gain runs use $N=8$ candidate paths with fixed $K=2$ tokens per step. We report task accuracy and sequence-level calibration. 

We compare the sequence scores in Eq.~\ref{eq:bayeser-sequence-confidence} with binary task correctness using 15 equal-width bins for ECE (Section~\ref{sec:background_calibration}) and report binary Brier scores. We compute scores over generated-content tokens and final-answer tokens, yielding ECE$_{\mathrm{gen}}$ and ECE$_{\mathrm{ans}}$. For molecule generation, ECE is evaluated against chemical validity, while quality yield measures the fraction of attempts meeting the QED and SA criteria in Table~\ref{tab:scaffold_decoration}.

\subsection{Main results}
\label{sec:results}

\subsubsection{Mathematical reasoning and programming}

Table~\ref{tab:main_results} shows that \method{} improves sequence-confidence calibration on several tasks, with task-dependent changes in accuracy. On MBPP, the accuracy-oriented setting approximately preserves threshold accuracy (57.53\% versus 57.60\%) while reducing generated-content ECE from 0.2918 to 0.2354. The calibration-oriented setting lowers ECE further to 0.1632 at 56.73\% accuracy. Transferring the MBPP posterior to HumanEval also reduces ECE and Brier scores without refitting: generated-content ECE falls from 0.3090 to 0.2016 in the calibration-oriented setting, with 55.08\% accuracy versus 54.88\% for threshold. \baseentropy{} retains the highest HumanEval accuracy at 57.32\%.

On GSM8K, \baseentropy{} achieves the highest accuracy (78.17\%) and lowest generated-content Brier score, while the calibration-oriented \method{} setting achieves the lowest ECE in both scopes. On MATH500, \method{} improves accuracy from 42.40\% to 43.53\% and reduces ECE and Brier scores. The accuracy--calibration tradeoff will be further elaborated in Section~\ref{sec:hyperparameter_tradeoffs}. 

\subsubsection{Planning}

\begin{table}[t]
\caption{Evaluation on planning tasks. }
\label{tab:planning_results}
\centering
\scriptsize
\setlength{\tabcolsep}{4pt}
\setlength{\extrarowheight}{1pt}
\newcommand{\resultdet}[2]{\ensuremath{#1\,(#2)}}
\newcommand{\resultseed}[3]{\ensuremath{#1_{\scriptscriptstyle #2}\,(#3)}}
\resizebox{\textwidth}{!}{%
\begin{tabular}{@{}llrrrr@{}}
\toprule
Dataset & Method & Accuracy (\%) $\uparrow$ & ECE$_{\mathrm{gen}}$ ($\times100$) $\downarrow$ & ECE$_{\mathrm{ans}}$ ($\times100$) $\downarrow$ & Brier$_{\mathrm{gen}}$ ($\times100$) $\downarrow$ \\
\midrule
\textbf{Sudoku} & Threshold (Fast-dllm) & $\mathbf{97.60}$ & 3.13 & 3.13 & $\mathbf{2.24}$ \\
 & \baseentropy & \resultdet{\mathbf{97.60}}{+0.00} & \resultdet{\underline{2.98}}{+4.79} & \resultdet{\underline{2.98}}{+4.79} & \resultdet{\underline{2.28}}{-1.54} \\
 & Info-gain & \resultseed{96.07}{0.50}{-1.57} & \resultseed{\mathbf{2.54}}{0.08}{+18.85} & \resultseed{\mathbf{2.54}}{0.08}{+18.85} & \resultseed{3.39}{0.37}{-51.35} \\
 & \method{} (best accuracy) & \resultseed{\underline{97.53}}{0.31}{-0.07} & \resultseed{3.96}{0.03}{-26.52} & \resultseed{3.96}{0.03}{-26.52} & \resultseed{2.40}{0.17}{-7.00} \\
 & \method{} (best calibration) & \resultseed{97.40}{0.20}{-0.20} & \resultseed{2.99}{0.27}{+4.47} & \resultseed{2.99}{0.27}{+4.47} & \resultseed{2.39}{0.16}{-6.56} \\
\midrule
\textbf{Countdown} & Threshold (Fast-dllm) & 18.90 & 66.46 & 61.54 & $58.56$ \\
 & \baseentropy & \resultdet{23.90}{+26.46} & \resultdet{60.07}{+9.61} & \resultdet{54.67}{+11.16} & \resultdet{53.21}{+9.14} \\
 & Info-gain & \resultseed{\mathbf{41.47}}{2.03}{+119.42} & \resultseed{45.63}{1.77}{+31.34} & \resultseed{43.56}{1.08}{+29.22} & \resultseed{42.40}{1.16}{+27.60} \\
 & \method{} (best accuracy) & \resultseed{\underline{32.13}}{0.67}{+70.00} & \resultseed{\underline{28.98}}{0.21}{+56.39} & \resultseed{\mathbf{27.92}}{0.68}{+54.63} & \resultseed{\underline{26.71}}{0.18}{+54.39} \\
 & \method{} (best calibration) & \resultseed{27.80}{0.78}{+47.09} & \resultseed{\mathbf{25.70}}{0.50}{+61.33} & \resultseed{\underline{30.24}}{0.81}{+50.86} & \resultseed{\mathbf{22.83}}{0.08}{+61.01} \\
\bottomrule
\end{tabular}%
}
\end{table}

Table~\ref{tab:planning_results} shows contrasting outcomes on the two planning tasks. Sudoku offers little advantage for Bayesian decoding: threshold and \baseentropy{} reach 97.60\% accuracy, and neither posterior setting improves on threshold's Brier score. Countdown exhibits the strongest accuracy--calibration trade-off. The accuracy-oriented \method{} setting reaches 32.13\% accuracy, exceeding threshold (18.90\%) and \baseentropy{} (23.90\%), but falling below Info-gain (41.47\%). Its generated-content ECE and Brier score are lower than Info-gain's: 0.2898 versus 0.4563 and 0.2671 versus 0.4240, respectively. The calibration-oriented setting further reduces generated-content ECE to 0.2570 at 27.80\% accuracy. 

\subsubsection{Molecular generation}

\begin{table}[t]
\centering
\caption{Evaluation on GenMol V2 Scaffold Decoration. Quality Yield measures the percentage of generated attempts satisfying both drug-likeness ($\text{QED} \ge 0.6$) and synthetic accessibility ($\text{SA} \le 4.0$). }
\label{tab:scaffold_decoration}
\resizebox{\textwidth}{!}{%
\begin{tabular}{lccccc}
\toprule
Method & Quality Yield (\%) $\uparrow$ & Frag. Valid (\%) $\uparrow$ & QED $\uparrow$ & SA $\downarrow$ & ECE $\downarrow$ \\
\midrule
Native GenMol & $11.78 \pm 0.69$ & $99.33 \pm 0.33$ & $0.6369 \pm 0.0104$ & $4.3932 \pm 0.0362$ & $45.08 \pm 0.29$ \\
\baseentropy{} & $\underline{14.89} \pm 0.69$ & $\underline{99.89} \pm 0.19$ & $0.6599 \pm 0.0011$ & $4.2809 \pm 0.0106$ & $35.40 \pm 0.59$ \\
Info-gain & $3.00 \pm 0.67$ & $\mathbf{100.00 \pm 0.00}$ & $\mathbf{0.6704 \pm 0.0206}$ & $\mathbf{4.1528 \pm 0.0595}$ & $\mathbf{24.56 \pm 1.71}$ \\
DPRM \citep{bu2026dprm} & $12.44 \pm 0.69$ & $\mathbf{100.00 \pm 0.00}$ & $\underline{0.6619} \pm 0.0117$ & $\underline{4.2682} \pm 0.0742$ & $\underline{30.62} \pm 1.33$ \\
\textbf{\method{} (Ours)} & $\mathbf{15.33 \pm 0.33}$ & $99.67 \pm 0.00$ & $0.6602 \pm 0.0006$ & $4.2900 \pm 0.0062$ & $35.36 \pm 0.73$ \\
\bottomrule
\end{tabular}%
}
\end{table}

On GenMol V2 scaffold decoration (Table~\ref{tab:scaffold_decoration}), \method{} achieves the highest mean quality yield at 15.33\%, compared with 11.78\% for native GenMol and 14.89\% for \baseentropy{}. Its ECE is lower than native GenMol's (35.36 versus 45.08 on the table's scale), but nearly identical to \baseentropy{}'s (35.40). These metrics capture different aspects of generation: ECE concerns chemical validity, whereas quality yield requires both drug-likeness and synthetic accessibility.

\subsection{Hyperparameter sensitivity and accuracy--calibration trade-offs}
\label{sec:hyperparameter_tradeoffs}

\begin{figure}[t]
\centering
\includegraphics[width=0.95\linewidth]{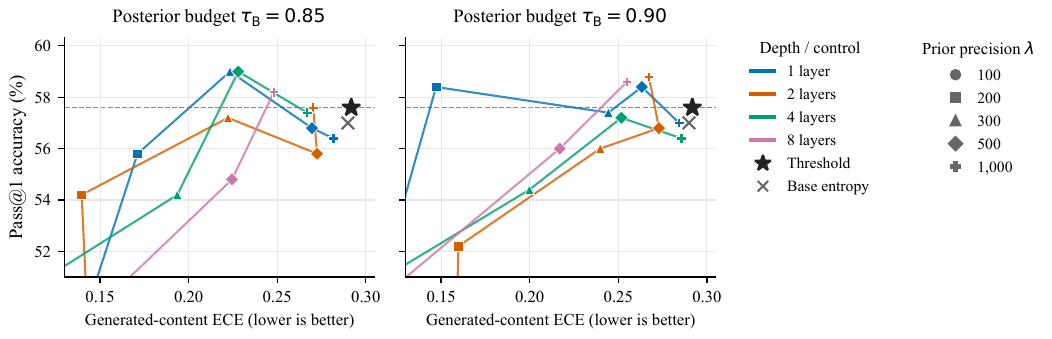} \caption{Accuracy--calibration trade-off for MBPP with different hyperparameter choices. }
\label{fig:mbpp_accuracy_ece}
\end{figure}

We additionally evaluate how prior precision $\lambda$, the number of uncertain trailing layers $L_s$, and the commitment budget jointly affect the accuracy--calibration tradeoff. In Figure~\ref{fig:mbpp_accuracy_ece}, each panel shows five prior precisions ($\lambda \in \{100, 200, 300, 500, 1000\}$) at four depths. Intermediate precision provides operating points with lower ECE and comparable or higher accuracy. Further weakening the prior can reduce ECE while substantially degrading accuracy.

Depth and commitment budget also interact. A single uncertain layer reaches 59.0\% at ECE 0.223 for $\tau_B=0.85$, $\lambda=300$, and 58.4\% at ECE 0.147 for $\tau_B=0.90$, $\lambda=200$. These examples show that a shallow posterior can offer a favorable accuracy--calibration trade-off and that additional uncertain layers do not consistently improve it. 

\subsection{Computational efficiency}
\label{sec:efficiency}

\begin{figure}[t]
\centering
\includegraphics[width=0.8\linewidth]{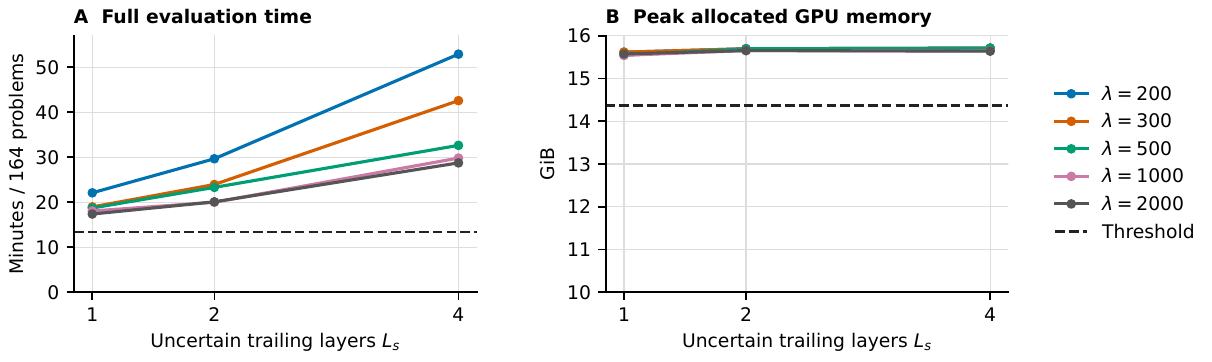}
\caption{Dream HumanEval cost versus uncertain depth.}
\label{fig:dream_humaneval_depth_cost}
\end{figure}

Figure~\ref{fig:dream_humaneval_depth_cost} shows HumanEval cost on A100 GPUs with $S=10$ samples and $\tau_{\mathrm{B}}=0.90$. Across tested priors, increasing uncertain depth from one to four layers raises evaluation time from 17.4--22.1 to 28.8--52.9 minutes, versus 13.4 minutes for threshold. Peak allocated memory remains within 15.54--15.71 GiB, versus 14.38 GiB for threshold. 
Deeper posteriors require longer sampled forward passes, while weaker priors can prolong decoding. Shallow posteriors therefore limit runtime overhead. Memory varies little with depth.

\section{Conclusion and Discussion}\label{sec:conclusion}
We introduced \method{}, a post-hoc Bayesian decoding scheme that uses posterior predictive uncertainty to guide position ordering, token selection, and commitment budgets in masked diffusion language models. 
Its curvature-only Laplace-LoRA construction preserves the released checkpoint at the posterior mean and requires no adapter fine-tuning. 
The experiments show that posterior prediction can improve sequence-confidence calibration while preserving or improving task performance at selected operating points. 
In future work, we plan to extend the single Gaussian posterior to multiple Gaussian posteriors across different adapter-parameter layers to better capture parameter uncertainties and enhance our entropy-based decoding scheme.

\newpage
\subsubsection*{Reproducibility Statement}
The pretrained checkpoints used in this work (Dream-v0-Instruct-7B,
LLaDA-8B-Instruct, and GenMol V2) and the MBPP, HumanEval, GSM8K, and MATH500
benchmarks are publicly available. Planning evaluations use fixed public Sudoku
and Countdown test sets; separate synthetic training sets are generated
deterministically with test inputs excluded. Molecular-generation experiments
use public GenMol fragments and calibration data drawn from the training split
of \texttt{datamol-io/safe-gpt}.
Section~\ref{sec:method} and Appendix~\ref{app:details} describe the method,
posterior construction, and predictive computation. Section~\ref{sec:experiments}
describes the benchmarks and metrics; Tables~\ref{tab:hyperparameters}
and~\ref{tab:selected_configs} specify the Dream configurations, and
Appendix~\ref{app:llada10} reports the additional LLaDA evaluation settings.
Code, posterior artifacts, data-preparation scripts, and evaluation configurations
will be open-sourced upon publication.

\subsubsection*{Use of AI Statement}
Generative AI tools and large language models were used to assist with code
implementation, finding related work, manuscript editing, grammar polishing,
and typesetting.
AI-assisted code was checked by the authors. The authors take responsibility
for the final manuscript, including its code, experimental results, and claims.

\subsubsection*{Ethics Statement}
This work studies uncertainty-aware decoding using pretrained models and public
benchmark datasets. Improved benchmark calibration does not guarantee reliable
predictions under distribution shift or safe deployment in high-stakes settings.
The molecular-generation experiments evaluate computational properties; the
reported metrics do not establish biological efficacy, toxicity, or safety.
Generated molecules would require independent expert assessment and experimental
validation before practical use.

\ifdefined\arxivacknowledgments
\arxivacknowledgments
\fi

\bibliography{references}

\begin{thebibliography}{26}
\providecommand{\natexlab}[1]{#1}
\providecommand{\url}[1]{\texttt{#1}}
\expandafter\ifx\csname urlstyle\endcsname\relax
  \providecommand{\doi}[1]{doi: #1}\else
  \providecommand{\doi}{doi: \begingroup \urlstyle{rm}\Url}\fi

\bibitem[Ben-Hamu et~al.(2025)Ben-Hamu, Gat, Severo, Nolte, and
  Karrer]{benhamu2025entropy}
Heli Ben-Hamu, Itai Gat, Daniel Severo, Niklas Nolte, and Brian Karrer.
\newblock Accelerated sampling from masked diffusion models via entropy bounded
  unmasking.
\newblock In \emph{Advances in Neural Information Processing Systems}, 2025.
\newblock URL \url{https://arxiv.org/abs/2505.24857}.

\bibitem[Brier(1950)]{brier1950verification}
Glenn~W. Brier.
\newblock Verification of forecasts expressed in terms of probability.
\newblock \emph{Monthly Weather Review}, 78\penalty0 (1):\penalty0 1--3, 1950.

\bibitem[Bu et~al.(2026)Bu, Huang, Han, Wu, Wong, Zhang, Suzuki, and
  Nitanda]{bu2026dprm}
Dake Bu, Wei Huang, Andi Han, Si~Wu, Hau-San Wong, Qingfu Zhang, Taiji Suzuki,
  and Atsushi Nitanda.
\newblock {DPRM}: A plug-in {Doob} $h$-transform-induced token-ordering module
  for discrete diffusion models, 2026.
\newblock URL \url{https://arxiv.org/abs/2604.24357}.

\bibitem[Chang et~al.(2022)Chang, Zhang, Jiang, Liu, and
  Freeman]{chang2022maskgit}
Huiwen Chang, Han Zhang, Lu~Jiang, Ce~Liu, and William~T. Freeman.
\newblock Maskgit: Masked generative image transformer.
\newblock In \emph{Proceedings of the IEEE/CVF Conference on Computer Vision
  and Pattern Recognition (CVPR)}, 2022.

\bibitem[Daxberger et~al.(2021)Daxberger, Kristiadi, Immer, Eschenhagen, Bauer,
  and Hennig]{daxberger2021laplace}
Erik Daxberger, Agustinus Kristiadi, Alexander Immer, Runa Eschenhagen,
  Matthias Bauer, and Philipp Hennig.
\newblock {L}aplace redux --- effortless {B}ayesian deep learning.
\newblock In \emph{Advances in Neural Information Processing Systems}, 2021.

\bibitem[Guo et~al.(2017)Guo, Pleiss, Sun, and Weinberger]{pmlr-v70-guo17a}
Chuan Guo, Geoff Pleiss, Yu~Sun, and Kilian~Q. Weinberger.
\newblock On calibration of modern neural networks.
\newblock In \emph{Proceedings of the 34th International Conference on Machine
  Learning}, volume~70 of \emph{Proceedings of Machine Learning Research}, pp.\
   1321--1330. PMLR, 2017.
\newblock URL \url{https://proceedings.mlr.press/v70/guo17a.html}.

\bibitem[Houlsby et~al.(2011)Houlsby, Husz{\'a}r, Ghahramani, and
  Lengyel]{houlsby2011bald}
Neil Houlsby, Ferenc Husz{\'a}r, Zoubin Ghahramani, and M{\'a}t{\'e} Lengyel.
\newblock {B}ayesian active learning for classification and preference
  learning, 2011.
\newblock URL \url{https://arxiv.org/abs/1112.5745}.

\bibitem[Hu et~al.(2021)Hu, Shen, Wallis, Allen-Zhu, Li, Wang, Wang, and
  Chen]{hu2021lora}
Edward~J. Hu, Yelong Shen, Phillip Wallis, Zeyuan Allen-Zhu, Yuanzhi Li, Shean
  Wang, Lu~Wang, and Weizhu Chen.
\newblock Lo{RA}: Low-rank adaptation of large language models, 2021.
\newblock URL \url{https://arxiv.org/abs/2106.09685}.

\bibitem[Jazbec et~al.(2025)Jazbec, Wong-Toi, Xia, Zhang, Nalisnick, and
  Mandt]{jazbec2025generative}
Metod Jazbec, Eliot Wong-Toi, Guoxuan Xia, Dan Zhang, Eric Nalisnick, and
  Stephan Mandt.
\newblock Generative uncertainty in diffusion models.
\newblock \emph{arXiv preprint arXiv:2502.20946}, 2025.
\newblock URL \url{https://arxiv.org/abs/2502.20946}.

\bibitem[Jazbec et~al.(2026)Jazbec, Olausson, B{\'e}thune, Ablin, Kirchhof,
  Monteiro, Turrisi, Ramapuram, and Cuturi]{jazbec2025learning}
Metod Jazbec, Theo~X. Olausson, Louis B{\'e}thune, Pierre Ablin, Michael
  Kirchhof, Jo{\~a}o Monteiro, Victor Turrisi, Jason Ramapuram, and Marco
  Cuturi.
\newblock Learning unmasking policies for diffusion language models.
\newblock In \emph{International Conference on Machine Learning}, 2026.
\newblock URL \url{https://arxiv.org/abs/2512.09106}.

\bibitem[Kou et~al.(2023)Kou, Gan, Wang, Li, and Deng]{kou2023bayesdiff}
Siqi Kou, Lei Gan, Dequan Wang, Chongxuan Li, and Zhijie Deng.
\newblock {B}ayes{D}iff: Estimating pixel-wise uncertainty in diffusion via
  {B}ayesian inference.
\newblock \emph{arXiv preprint arXiv:2310.11142}, 2023.
\newblock URL \url{https://arxiv.org/abs/2310.11142}.

\bibitem[Lee et~al.(2025)Lee, Kim, Park, and Park]{lee2025lookahead}
Sanghyun Lee, Seungryong Kim, Jongho Park, and Dongmin Park.
\newblock Lookahead unmasking elicits accurate decoding in diffusion language
  models, 2025.
\newblock URL \url{https://arxiv.org/abs/2511.05563}.

\bibitem[Li et~al.(2026)Li, Liu, and Zhang]{li2026confidence}
Yansi Li, Gongshen Liu, and Zhuosheng Zhang.
\newblock The confidence paradox: Unveiling the latent discriminative power of
  diffusion large language models in mathematical reasoning.
\newblock In Maria Liakata, Viviane~P. Moreira, Jiajun Zhang, and David Jurgens
  (eds.), \emph{Findings of the {A}ssociation for {C}omputational
  {L}inguistics: {ACL} 2026}, pp.\  43179--43196, San Diego, California, United
  States, July 2026. Association for Computational Linguistics.
\newblock ISBN 979-8-89176-395-1.
\newblock \doi{10.18653/v1/2026.findings-acl.2142}.
\newblock URL \url{https://aclanthology.org/2026.findings-acl.2142/}.

\bibitem[MacKay(1992)]{mackay1992laplace}
David J.~C. MacKay.
\newblock A practical {B}ayesian framework for backpropagation networks.
\newblock \emph{Neural Computation}, 4\penalty0 (3):\penalty0 448--472, 1992.

\bibitem[Martens \& Grosse(2015)Martens and Grosse]{martens2015kfac}
James Martens and Roger Grosse.
\newblock Optimizing neural networks with {K}ronecker-factored approximate
  curvature.
\newblock In \emph{International Conference on Machine Learning}, pp.\
  2408--2417, 2015.

\bibitem[Naeini et~al.(2015)Naeini, Cooper, and
  Hauskrecht]{naeini2015obtaining}
Mahdi~Pakdaman Naeini, Gregory~F. Cooper, and Milos Hauskrecht.
\newblock Obtaining well calibrated probabilities using bayesian binning.
\newblock In \emph{Proceedings of the Twenty-Ninth AAAI Conference on
  Artificial Intelligence}, pp.\  2901--2907, 2015.
\newblock \doi{10.1609/aaai.v29i1.9602}.

\bibitem[Nie et~al.(2025)Nie, Zhu, You, Zhang, Ou, Hu, Zhou, Lin, Wen, and
  Li]{nie2025llada}
Shen Nie, Fengqi Zhu, Zebin You, Xiaolu Zhang, Jingyang Ou, Jun Hu, Jun Zhou,
  Yankai Lin, Ji-Rong Wen, and Chongxuan Li.
\newblock Large language diffusion models.
\newblock \emph{arXiv preprint arXiv:2502.09992}, 2025.
\newblock URL \url{https://arxiv.org/abs/2502.09992}.

\bibitem[Ritter et~al.(2018)Ritter, Botev, and Barber]{ritter2018scalable}
Hippolyt Ritter, Aleksandar Botev, and David Barber.
\newblock A scalable {L}aplace approximation for neural networks.
\newblock In \emph{International Conference on Learning Representations}, 2018.

\bibitem[Sahoo et~al.(2024)Sahoo, Arriola, Schiff, Gokaslan, Marroquin, Chiu,
  Rush, and Kuleshov]{sahoo2024mdlm}
Subham~Sekhar Sahoo, Marianne Arriola, Yair Schiff, Aaron Gokaslan, Edgar
  Marroquin, Justin~T Chiu, Alexander Rush, and Volodymyr Kuleshov.
\newblock Simple and effective masked diffusion language models.
\newblock In \emph{Advances in Neural Information Processing Systems}, 2024.

\bibitem[Shi et~al.(2025)Shi, Wang, Han, Zhang, and Wang]{shi2024tfb}
Haizhou Shi, Yibin Wang, Ligong Han, Huan Zhang, and Hao Wang.
\newblock Training-free {B}ayesianization for low-rank adapters of large
  language models.
\newblock In \emph{Advances in Neural Information Processing Systems}, 2025.
\newblock URL \url{https://arxiv.org/abs/2412.05723}.

\bibitem[Shu et~al.(2026)Shu, Tian, Xu, Wang, and Chen]{shu2026dcd}
Yingte Shu, Yuchuan Tian, Chao Xu, Yunhe Wang, and Hanting Chen.
\newblock Deferred commitment decoding for diffusion language models with
  confidence-aware sliding windows.
\newblock \emph{arXiv preprint arXiv:2601.02076}, 2026.
\newblock URL \url{https://arxiv.org/abs/2601.02076}.

\bibitem[Wang et~al.(2024)Wang, Shi, Han, Metaxas, and Wang]{wang2024blob}
Yibin Wang, Haizhou Shi, Ligong Han, Dimitris Metaxas, and Hao Wang.
\newblock {BL}o{B}: {B}ayesian low-rank adaptation by backpropagation for large
  language models.
\newblock In \emph{Advances in Neural Information Processing Systems},
  volume~37, pp.\  67758--67794, 2024.

\bibitem[Wu et~al.(2025)Wu, Zhang, Xue, Liu, Diao, Zhu, Luo, Han, and
  Xie]{wu2025fastdllm}
Chengyue Wu, Hao Zhang, Shuchen Xue, Zhijian Liu, Shizhe Diao, Ligeng Zhu, Ping
  Luo, Song Han, and Enze Xie.
\newblock Fast-d{LLM}: Training-free acceleration of diffusion {LLM} by
  enabling {KV} cache and parallel decoding.
\newblock \emph{arXiv preprint arXiv:2505.22618}, 2025.
\newblock URL \url{https://arxiv.org/abs/2505.22618}.

\bibitem[Yang et~al.(2024)Yang, Robeyns, Wang, and
  Aitchison]{yang2024laplacelora}
Adam~X. Yang, Maxime Robeyns, Xi~Wang, and Laurence Aitchison.
\newblock {B}ayesian low-rank adaptation for large language models.
\newblock In \emph{International Conference on Learning Representations}, 2024.
\newblock URL \url{https://openreview.net/forum?id=FJiUyzOF1m}.

\bibitem[Yang et~al.(2026)Yang, Teoh, Yang, Zhang, and Lamb]{yang2026infogain}
Kaisen Yang, Jayden Teoh, Kaicheng Yang, Yitong Zhang, and Alex Lamb.
\newblock Improving sampling for masked diffusion models via information gain,
  2026.
\newblock URL \url{https://arxiv.org/abs/2602.18176}.

\bibitem[Ye et~al.(2025)Ye, Xie, Zheng, Gao, Wu, Jiang, Li, and
  Kong]{ye2025dream}
Jiacheng Ye, Zhihui Xie, Lin Zheng, Jiahui Gao, Zirui Wu, Xin Jiang, Zhenguo
  Li, and Lingpeng Kong.
\newblock Dream 7b: Diffusion large language models.
\newblock \emph{arXiv preprint arXiv:2508.15487}, 2025.
\newblock URL \url{https://arxiv.org/abs/2508.15487}.

\end{thebibliography}
\bibliographystyle{iclr2027_conference}

\newpage
\appendix

\section{Posterior and implementation details}
\label{app:details}

\subsection{Posterior fitting, sampling and computation}
\label{app:posterior_sampling}

We use the notation of Section~\ref{sec:method-posterior}: $\vtheta_0$ denotes the fixed pretrained parameters, $\vphi$ collects the LoRA adapter parameters, $j$ indexes selected transformer layers, and $m$ indexes individual adapter-matrix blocks. We fit model-Fisher curvature at $\vphi_0$, where the LoRA update is zero, using task-matched calibration prompts and suppressing the mask-token logit before softmax. Algorithm~\ref{alg:lora_fit} summarizes the procedure. For
posterior draws we use the exact factorization implied by
Eq.~\ref{eq:bayeser-posterior}. Let $\vphi_m$ be the vectorized adapter matrix for block $m$, with mean $\vphi_{0,m}$; here $\operatorname{vec}$ stacks rows, consistent with the ordering $G_m\otimes C_m$. Draw a matrix $z_m^{(s)}$ of independent standard normal entries in the shape of that adapter matrix and expand it in the retained basis,
$z_m^{(s)}=\sum_{a,b}\xi_{m,ab}^{(s)}u^G_{m,a}(u^C_{m,b})^\top+z_{m,\perp}^{(s)}$. The adapter draw is
\begin{equation}
    \vphi_m^{(s)}
    =\vphi_{0,m}
      +\operatorname{vec}\!\left[
      \sum_{a,b}\frac{\xi_{m,ab}^{(s)}}
        {\sqrt{\lambda+N_m g_{m,a}c_{m,b}}}\,u^G_{m,a}(u^C_{m,b})^\top
      +\frac{1}{\sqrt{\lambda}}z_{m,\perp}^{(s)}\right],
    \label{eq:weight_sampling}
\end{equation}
where $c_{m,b}$ and $g_{m,a}$ are the retained eigenvalues of $C_m$ and $G_m$
and $N_m$ is the represented source-position observation count defined in the main text. The coefficients $\xi_{m,ab}^{(s)}$ are independent standard normal draws. The retained-basis perturbation is the data- and
prior-informed component; the last term is the prior-only
complement corresponding to truncated directions. Each block draws its own
independent $z_m^{(s)}$, realizing the block-independence assumption above.
Substituting Eq.~\ref{eq:weight_sampling} into
Eq.~\ref{eq:bayeser-lora} gives the nonlinear weight-space predictive
$p_{\vtheta_0,\vphi^{(s)}}^{(l)}(\cdot\mid\vx_t^{(1:L)})$ at position $l$. Normalizing the sampled logits on $\mathcal{V}_t^{(l)}$ gives $p_{t,s}^{(l)}$ from Section~\ref{sec:method-entropy}.

\begin{algorithm}[H]
\caption{Curvature-only KFAC Laplace-LoRA posterior}
\label{alg:lora_fit}
\begin{algorithmic}[1]
\REQUIRE Released model $f_{\vtheta_0}$; calibration set
$\mathcal{D}_{\mathrm{cal}}$; selected projections; LoRA rank $r$ and scale
$\alpha$; KFAC rank $k$; prior precision $\lambda$; per-block row-sampling cap $N_s$
\STATE Wrap each selected projection as
$\mW_j(\vphi_j)=\mW_{j,0}+\frac{\alpha}{r}\mB_j\mA_j$ with $\mB_{j,0}=0$
(Eq.~\ref{eq:bayeser-lora}); let $\vphi_0$ collect $\{\mA_{j,0},\mB_{j,0}\}_j$, with projection indices omitted as in the main text
\STATE Initialize weighted sums $S^C_m=0$, $S^G_m=0$ and counts $N_m=0$ for each block $m$
\FOR{each calibration batch of masked sequences $\widetilde{\mathbf{x}}_n$ and target masks}
    \STATE Forward the masked sequence; suppress the mask-token logit before softmax and sample one categorical target from the model distribution at each masked target position
    \STATE Backpropagate the summed cross-entropy against these sampled targets
    \FOR{each adapter-matrix block $m$ ($\mA_j$ and $\mB_j$ separately)}
        \STATE Collect paired rows $(h_{m,n},g_{m,n})$ at positions with nonzero gradient; skip blocks with no eligible rows
        \STATE Uniformly subsample up to $N_s$ rows; set each retained row's weight $w_n$ to the number of eligible rows divided by the number retained
        \STATE $S^C_m\leftarrow S^C_m+\sum_n w_n h_{m,n}h_{m,n}^\top$, $S^G_m\leftarrow S^G_m+\sum_n w_n g_{m,n}g_{m,n}^\top$, and $N_m\leftarrow N_m+\sum_n w_n$
    \ENDFOR
\ENDFOR
\FOR{each block $m$}
    \IF{$N_m=0$}
        \STATE Set $F_m=0$ (prior-only block)
    \ELSE
    \STATE Normalize $C_m=S^C_m/N_m$ and $G_m=S^G_m/N_m$
    \STATE Eigendecompose $C_m$ and $G_m$; retain the top-$k$ directions of
    large factors
    \STATE Store $(\{u^C_{m,b}\},\{c_{m,b}\},\{u^G_{m,a}\},\{g_{m,a}\},N_m)$
    with curvature $F_m=N_m(G_m\otimes C_m)$
    \ENDIF
\ENDFOR
\STATE Form $F_{\mathrm{KFAC}}=\operatorname{blockdiag}(F_1,\ldots,F_R)$ and
$q_\lambda(\vphi)=\mathcal{N}(\vphi_0,(F_{\mathrm{KFAC}}+\lambda I)^{-1})$
\RETURN $q_\lambda$ and the per-block eigenbasis; $\lambda$ is applied at inference
\end{algorithmic}
\end{algorithm}

All reported evaluations are cache-free. Within each decoding step, activations
before the first uncertain layer are computed once and reused across posterior
draws; each draw evaluates the remaining layers with its sampled weights.
This reuse is exact because the preceding layers are unchanged across draws.
Activations are recomputed at every decoding step, and draws are batched where
memory permits.

\subsection{Execution protocols}
\label{app:protocol_details}

All rows in Tables~\ref{tab:main_results} and~\ref{tab:planning_results} use exact cache-free forward passes. Planning evaluations use the task geometry described in Section~\ref{sec:experiments}.

\paragraph{Task-specific decoding geometry.}
For the Dream experiments, MBPP, HumanEval, GSM8K, and MATH500
use sliding-window decoding with an initial window width of
16 tokens and a maximum width of 128 tokens. Sudoku uses
in-place infilling, whereas Countdown uses a 48-token block.
These planning-specific settings replace the sliding-window
configuration used for code generation and mathematical reasoning.

\begin{table}[H]
\caption{Main experimental configuration for the six-benchmark Dream study. HumanEval
uses the MBPP-trained posterior without refitting; all other posteriors are fitted on
task-specific or synthetic training data.}
\label{tab:hyperparameters}
\centering
\small
\begin{tabular}{ll}
\toprule
Component & Setting \\
\midrule
Mean model & Dream-v0-Instruct-7B \\
Benchmarks & MBPP, HumanEval, GSM8K, MATH500, Sudoku, Countdown \\
Prompt protocol & task-matched Dream instruction prompts \\
Posterior mean & released checkpoint; zero LoRA update \\
Curvature & categorical model Fisher, low-rank KFAC ($k=10$) \\
Target modules & trailing-layer query/value projections \\
Posterior draws & $S=10$ \\
Vocabulary support & $K_v=256$ \\
Predictive backend & nonlinear posterior weight sampling \\
Commit threshold & $\tau=0.9$; posterior budget gate $\tau_{\mathrm{B}}=0.85$ or $0.9$ \\
Ranking & posterior predictive entropy \\
Commit budget & posterior confidence gate for all selected \method{} rows \\
Window (code/math) & Sliding, initial 16 tokens, maximum 128 tokens \\
Geometry (Sudoku) & In-place infilling \\
Geometry (Countdown) & 48-token block \\
Token proposal & posterior argmax for all selected \method{} rows \\
Sequence confidence & geometric mean of commit-time probabilities, generated and answer scopes \\
ECE & 15 equal-width bins \\
\bottomrule
\end{tabular}
\end{table}

\begin{table}[H]
\caption{Hyperparameters for the evaluation in the main results. HumanEval reuses the MBPP-trained posterior; the remaining rows are task-fitted. }
\label{tab:selected_configs}
\centering
\small
\begin{tabular}{@{}llrrrrr@{}}
\toprule
Task & Target & $L_s$ & Rank & Alpha & Prior & $\tau_{\mathrm{B}}$ \\
\midrule
MBPP & Accuracy & 1 & 8 & 16 & 300 & 0.85 \\
MBPP & Calibration & 1 & 8 & 16 & 200 & 0.90 \\
HumanEval & Accuracy & 1 & 8 & 16 & 300 & 0.85 \\
HumanEval & Calibration & 1 & 8 & 16 & 200 & 0.90 \\
GSM8K & Accuracy & 4 & 8 & 16 & 2{,}000 & 0.90 \\
GSM8K & Calibration & 1 & 8 & 16 & 500 & 0.85 \\
MATH500 & Accuracy / calibration & 2 & 8 & 16 & 2{,}000 & 0.90 \\
Sudoku & Accuracy & 2 & 8 & 16 & 500 & 0.90 \\
Sudoku & Calibration & 1 & 8 & 16 & 1{,}000 & 0.90 \\
Countdown & Accuracy & 4 & 8 & 16 & 200 & 0.85 \\
Countdown & Calibration & 1 & 8 & 16 & 100 & 0.90 \\
\bottomrule
\end{tabular}
\end{table}

\clearpage
\section{Additional Dream accuracy--calibration sweeps}
\label{app:dream_sweeps}

\begin{figure}[htbp]
\centering
\includegraphics[width=\linewidth]{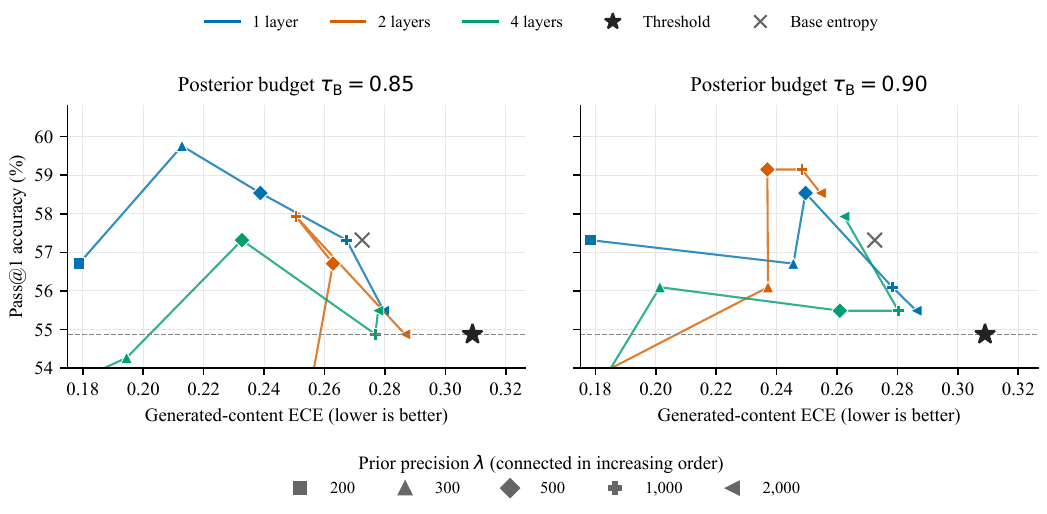}
\caption{HumanEval accuracy versus calibration.}
\label{fig:humaneval_accuracy_ece}
\end{figure}

\begin{figure}[htbp]
\centering
\includegraphics[width=\linewidth]{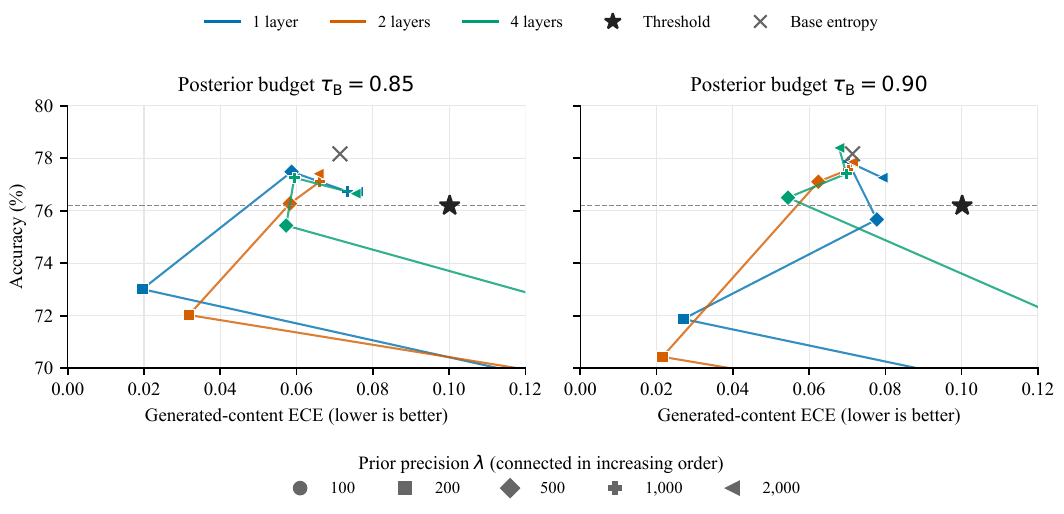}
\caption{GSM8K accuracy versus calibration.}
\label{fig:gsm8k_accuracy_ece}
\end{figure}

\begin{figure}[htbp]
\centering
\includegraphics[width=\linewidth]{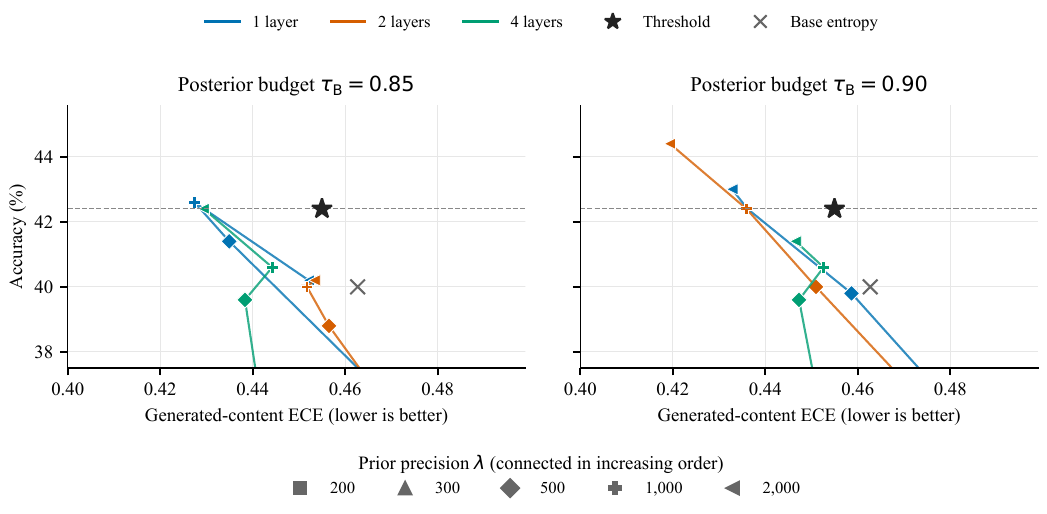}
\caption{MATH500 accuracy versus calibration.}
\label{fig:math500_accuracy_ece}
\end{figure}

\begin{figure}[htbp]
\centering
\includegraphics[width=\linewidth]{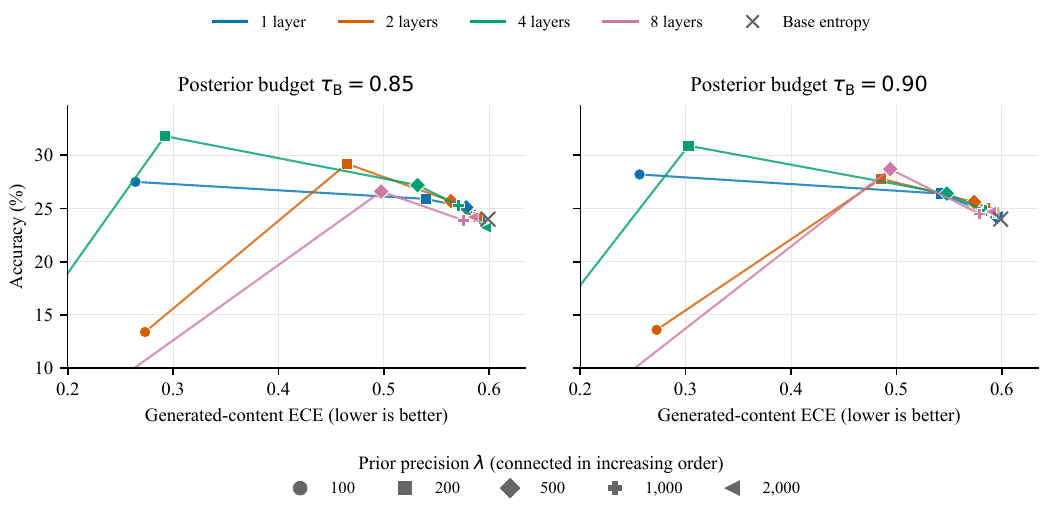}
\caption{Countdown accuracy versus calibration.}
\label{fig:countdown_accuracy_ece}
\end{figure}

Figures~\ref{fig:humaneval_accuracy_ece}--\ref{fig:countdown_accuracy_ece}
extend Figure~\ref{fig:mbpp_accuracy_ece} to four additional benchmarks.

\clearpage

\clearpage

\section{Additional measured cost}
\label{app:cost}

\begin{figure}[h]
\centering
\includegraphics[width=\linewidth]{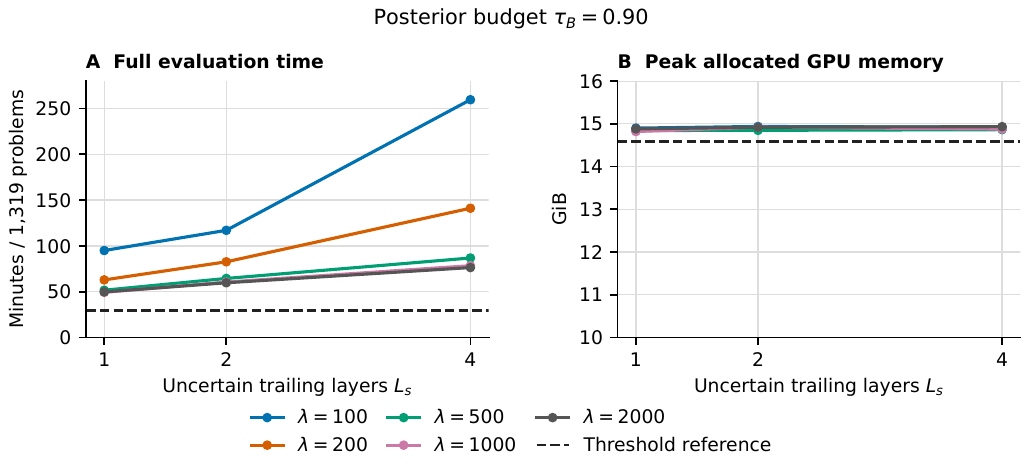}
\caption{\textbf{Measured Dream GSM8K cost versus uncertain depth.}
Full test set ($n=1{,}319$), GSM8K-fitted rank-8 q/v Laplace-LoRA with $S=10$, on H100 GPUs. Color denotes prior precision $\lambda$. Dashed lines show the confidence-threshold run. Wall time includes loading and scoring;
the loose prior increases decoding steps, especially at four layers.}
\label{fig:dream_gsm8k_depth_cost}
\end{figure}

Figure ~\ref{fig:dream_gsm8k_depth_cost} compares \method{} costs with Confidence-thresholding baseline on H100 GPU.

\clearpage
\raggedbottom
\section{LLaDA results}
\label{app:llada10}

Table~\ref{tab:llada10_results} reports full-test results for \texttt{GSAI-ML/LLaDA-8B-Instruct}. The recorded MBPP, GSM8K and MATH500 curvature-only sweeps use rank-8 q/v adapters ($\alpha=16$), KFAC rank 10, and vocabulary support $K_v=256$. HumanEval retains the MBPP transfer configuration described above. Info-gain uses $N=8$, fixed $K=2$.

\begin{table}[H]
\caption{LLaDA full-test results. Accuracy is reported in percent; ECE and generated-content Brier values and their standard deviations are multiplied by 100. Subscripts are the standard deviations of three seeds on the corresponding column scale. Numbers in parentheses are relative percentage improvements over confidence-thresholding decoding. Bold values indicate the best result and underlined values indicate the second-best result per metric within each benchmark, including ties.}
\label{tab:llada10_results}
\centering
\scriptsize
\setlength{\tabcolsep}{4pt}
\setlength{\extrarowheight}{1pt}
\newcommand{\lladaresult}[2]{\ensuremath{#1\,(#2)}}
\newcommand{\lladaseed}[3]{\ensuremath{#1_{\scriptscriptstyle #2}\,(#3)}}
\resizebox{\textwidth}{!}{%
\begin{tabular}{@{}lrrrr@{}}
\toprule
Method & Accuracy (\%) $\uparrow$ & ECE$_{\mathrm{gen}}$ ($\times100$) $\downarrow$ & ECE$_{\mathrm{ans}}$ ($\times100$) $\downarrow$ & Brier$_{\mathrm{gen}}$ ($\times100$) $\downarrow$ \\
\midrule
\multicolumn{5}{@{}l}{\textbf{MBPP (500)}} \\
Threshold (Fast-dllm) & $39.40$ & $47.63$ & $50.01$ & $46.14$ \\
\baseentropy & \lladaresult{41.20}{+4.57} & \lladaresult{\underline{44.51}}{+6.53} & \lladaresult{\underline{46.84}}{+6.33} & \lladaresult{\underline{43.48}}{+5.76} \\
Info-gain & \lladaseed{\underline{41.27}}{0.61}{+4.74} & \lladaseed{44.96}{0.50}{+5.59} & \lladaseed{48.01}{0.52}{+4.00} & \lladaseed{43.52}{0.26}{+5.67} \\
\method{} & \lladaseed{\mathbf{41.33}}{0.58}{+4.91} & \lladaseed{\mathbf{43.57}}{0.67}{+8.52} & \lladaseed{\mathbf{45.89}}{0.66}{+8.23} & \lladaseed{\mathbf{42.70}}{0.45}{+7.46} \\
\midrule
\multicolumn{5}{@{}l}{\textbf{HumanEval (164)}} \\
Threshold (Fast-dllm) & $18.29$ & $69.05$ & $66.50$ & $62.26$ \\
\baseentropy & \lladaresult{\underline{22.56}}{+23.33} & \lladaresult{\underline{63.34}}{+8.26} & \lladaresult{\underline{60.48}}{+9.04} & \lladaresult{\underline{57.46}}{+7.70} \\
Info-gain & \lladaseed{15.65}{2.31}{-14.44} & \lladaseed{70.51}{2.11}{-2.11} & \lladaseed{68.38}{1.89}{-2.84} & \lladaseed{62.95}{1.39}{-1.11} \\
\method{} & \lladaseed{\mathbf{23.17}}{0.00}{+26.67} & \lladaseed{\mathbf{62.76}}{0.02}{+9.11} & \lladaseed{\mathbf{59.81}}{0.03}{+10.06} & \lladaseed{\mathbf{56.99}}{0.07}{+8.46} \\
\midrule
\multicolumn{5}{@{}l}{\textbf{GSM8K (1{,}319)}} \\
Threshold (Fast-dllm) & $\underline{76.12}$ & $17.49$ & $\underline{20.59}$ & $20.63$ \\
\baseentropy & \lladaresult{75.51}{-0.80} & \lladaresult{17.11}{+2.21} & \lladaresult{20.98}{-1.89} & \lladaresult{20.64}{-0.05} \\
Info-gain & \lladaseed{\mathbf{78.09}}{0.53}{+2.59} & \lladaseed{\mathbf{15.87}}{0.50}{+9.27} & \lladaseed{\mathbf{20.21}}{0.58}{+1.87} & \lladaseed{\mathbf{18.88}}{0.45}{+8.53} \\
\method{} & \lladaseed{75.92}{0.37}{-0.27} & \lladaseed{\underline{16.67}}{0.37}{+4.68} & \lladaseed{20.64}{0.40}{-0.21} & \lladaseed{\underline{20.24}}{0.30}{+1.89} \\
\midrule
\multicolumn{5}{@{}l}{\textbf{MATH500 (500)}} \\
Threshold (Fast-dllm) & $\underline{38.00}$ & $\underline{53.68}$ & $\underline{55.23}$ & $51.44$ \\
\baseentropy & \lladaresult{\mathbf{38.20}}{+0.53} & \lladaresult{\mathbf{52.13}}{+2.90} & \lladaresult{\mathbf{55.01}}{+0.41} & \lladaresult{\mathbf{49.67}}{+3.43} \\
Info-gain & \lladaseed{37.73}{0.42}{-0.70} & \lladaseed{53.81}{0.43}{-0.23} & \lladaseed{57.35}{0.60}{-3.84} & \lladaseed{\underline{51.08}}{0.41}{+0.69} \\
\method{} & \lladaseed{35.40}{0.72}{-6.84} & \lladaseed{54.60}{0.75}{-1.71} & \lladaseed{58.35}{0.57}{-5.64} & \lladaseed{51.48}{0.64}{-0.08} \\
\midrule
\multicolumn{5}{@{}l}{\textbf{Sudoku (500)}} \\
Threshold (Fast-dllm) & $\mathbf{85.60}$ & $\mathbf{9.57}$ & $\mathbf{9.57}$ & $\underline{10.42}$ \\
\baseentropy & \lladaresult{81.80}{-4.44} & \lladaresult{12.85}{-34.35} & \lladaresult{12.85}{-34.35} & \lladaresult{11.99}{-15.05} \\
Info-gain & \lladaseed{78.27}{1.10}{-8.57} & \lladaseed{10.10}{1.14}{-5.54} & \lladaseed{10.10}{1.14}{-5.54} & \lladaseed{14.85}{0.66}{-42.48} \\
\method{} & \lladaseed{\underline{85.47}}{1.47}{-0.16} & \lladaseed{\underline{9.81}}{0.24}{-2.49} & \lladaseed{\underline{9.81}}{0.24}{-2.49} & \lladaseed{\mathbf{10.26}}{0.82}{+1.57} \\
\midrule
\multicolumn{5}{@{}l}{\textbf{Countdown (1{,}000)}} \\
Threshold (Fast-dllm) & $\underline{35.10}$ & $\underline{53.65}$ & $52.14$ & $48.39$ \\
\baseentropy & \lladaresult{29.70}{-15.38} & \lladaresult{54.08}{-0.80} & \lladaresult{\underline{50.65}}{+2.85} & \lladaresult{47.92}{+0.97} \\
Info-gain & \lladaseed{\mathbf{39.57}}{1.55}{+12.73} & \lladaseed{\mathbf{49.28}}{1.23}{+8.14} & \lladaseed{\mathbf{46.99}}{1.05}{+9.88} & \lladaseed{\mathbf{45.38}}{1.04}{+6.22} \\
\method{} & \lladaseed{28.83}{0.38}{-17.85} & \lladaseed{54.07}{0.33}{-0.78} & \lladaseed{51.02}{0.36}{+2.14} & \lladaseed{\underline{47.68}}{0.20}{+1.46} \\
\bottomrule
\end{tabular}%
}
\end{table}

\clearpage
\section{Mutual-information ranking and Bayesian entropy decoding}
\label{app:mi_entropy_grid}

We compare \mirank{} with \method{} on the first 128 examples of MBPP and GSM8K.
The grid uses $L_s\in\{1,2,4\}$,
$\lambda\in\{100,200,500,1000,2000\}$, and
$\tau_{\mathrm B}\in\{0.85,0.90\}$ for \method{}.

\begin{table}[h]
\centering
\caption{Comparison between \mirank{} and \method{} on 128 examples of MBPP.
Accuracy is in percent; both ECE columns are multiplied by 100.}
\label{tab:mi_entropy_mbpp}
\scriptsize
\setlength{\tabcolsep}{2.5pt}
\resizebox{\textwidth}{!}{%
\begin{tabular}{@{}rr rrrr rrrr rrrr@{}}
\toprule
& & \multicolumn{4}{c}{\mirank{}} & \multicolumn{4}{c}{\method{}, $\tau_{\mathrm B}=0.85$} & \multicolumn{4}{c}{\method{}, $\tau_{\mathrm B}=0.90$} \\
\cmidrule(lr){3-6}\cmidrule(lr){7-10}\cmidrule(l){11-14}
$L_s$ & $\lambda$ & Acc. & ECE$_g$ & ECE$_a$ & Steps & Acc. & ECE$_g$ & ECE$_a$ & Steps & Acc. & ECE$_g$ & ECE$_a$ & Steps \\
\midrule
1 & 100 & 28.91 & 10.89 & 10.77 & 25.74 & 34.38 & 8.40 & 6.61 & 100.70 & \textbf{40.63} & 12.62 & 10.44 & 93.91 \\
1 & 200 & 48.44 & 19.63 & 21.55 & 38.02 & \textbf{57.03} & 16.42 & 18.14 & 59.38 & 55.47 & 19.27 & 20.04 & 67.27 \\
1 & 500 & 54.69 & 27.78 & 30.58 & 39.11 & \textbf{57.81} & 25.77 & 27.89 & 43.26 & \textbf{57.81} & 27.02 & 29.19 & 45.13 \\
1 & 1000 & 53.13 & 30.62 & 33.09 & 43.25 & 57.81 & 27.18 & 29.16 & 37.36 & \textbf{58.59} & 27.30 & 29.60 & 42.54 \\
1 & 2000 & \textbf{57.81} & 26.34 & 29.27 & 40.44 & \textbf{57.81} & 27.56 & 29.50 & 37.62 & 57.03 & 29.37 & 31.60 & 42.23 \\
\midrule
2 & 100 & \textbf{28.13} & 16.87 & 17.83 & 20.91 & 9.38 & 14.20 & 16.24 & 123.17 & 5.47 & 18.04 & 22.09 & 123.62 \\
2 & 200 & 45.31 & 15.05 & 16.11 & 33.48 & \textbf{55.47} & 13.03 & 14.74 & 72.17 & 53.13 & 15.10 & 17.64 & 76.32 \\
2 & 500 & 52.34 & 29.39 & 32.08 & 42.05 & \textbf{60.94} & 22.64 & 24.62 & 42.09 & 56.25 & 27.86 & 29.85 & 46.04 \\
2 & 1000 & 54.69 & 29.13 & 32.01 & 39.08 & \textbf{57.81} & 27.10 & 29.00 & 37.65 & 57.03 & 28.62 & 30.97 & 43.27 \\
2 & 2000 & 53.13 & 30.67 & 32.96 & 40.84 & \textbf{59.38} & 25.93 & 27.90 & 39.11 & 57.81 & 28.49 & 30.61 & 42.89 \\
\midrule
4 & 100 & \textbf{28.91} & 26.44 & 27.72 & 21.88 & 0.00 & 38.72 & 38.58 & 224.40 & 0.00 & 37.81 & 37.89 & 222.68 \\
4 & 200 & 42.19 & 12.85 & 11.29 & 27.82 & \textbf{46.09} & 5.04 & 10.20 & 112.18 & \textbf{46.09} & 6.27 & 10.23 & 119.48 \\
4 & 500 & 53.13 & 26.61 & 28.95 & 40.52 & \textbf{58.59} & 23.36 & 25.51 & 46.13 & 56.25 & 26.04 & 28.29 & 48.66 \\
4 & 1000 & 58.59 & 24.67 & 27.11 & 41.99 & \textbf{59.38} & 25.22 & 27.17 & 39.10 & 57.81 & 27.50 & 29.84 & 44.98 \\
4 & 2000 & 57.81 & 26.09 & 28.48 & 40.41 & \textbf{58.59} & 26.87 & 28.74 & 37.05 & \textbf{58.59} & 27.65 & 29.80 & 42.36 \\
\bottomrule
\end{tabular}%
}
\end{table}

\begin{table}[h]
\centering
\caption{Comparison between \mirank{} and \method{} on 128 examples of GSM8K.
Accuracy is in percent; both ECE columns are multiplied by 100.}
\label{tab:mi_entropy_gsm8k}
\scriptsize
\setlength{\tabcolsep}{2.5pt}
\resizebox{\textwidth}{!}{%
\begin{tabular}{@{}rr rrrr rrrr rrrr@{}}
\toprule
& & \multicolumn{4}{c}{\mirank{}} & \multicolumn{4}{c}{\method{}, $\tau_{\mathrm B}=0.85$} & \multicolumn{4}{c}{\method{}, $\tau_{\mathrm B}=0.90$} \\
\cmidrule(lr){3-6}\cmidrule(lr){7-10}\cmidrule(l){11-14}
$L_s$ & $\lambda$ & Acc. & ECE$_g$ & ECE$_a$ & Steps & Acc. & ECE$_g$ & ECE$_a$ & Steps & Acc. & ECE$_g$ & ECE$_a$ & Steps \\
\midrule
1 & 100 & 59.38 & 31.56 & 12.84 & 30.39 & 58.59 & 31.50 & 16.28 & 102.63 & \textbf{63.28} & 38.08 & 15.94 & 118.50 \\
1 & 200 & 64.84 & 10.14 & 19.34 & 41.83 & \textbf{77.34} & 5.56 & 15.76 & 70.57 & 75.78 & 6.65 & 16.95 & 77.88 \\
1 & 500 & 78.91 & 4.61 & 18.37 & 49.34 & \textbf{79.69} & 8.37 & 13.51 & 56.29 & 75.78 & 9.10 & 17.81 & 63.16 \\
1 & 1000 & 77.34 & 8.21 & 19.78 & 52.70 & 76.56 & 10.08 & 18.80 & 53.72 & \textbf{78.13} & 7.45 & 16.38 & 60.42 \\
1 & 2000 & 77.34 & 9.39 & 17.38 & 51.38 & 76.56 & 9.09 & 18.56 & 53.38 & \textbf{78.13} & 7.63 & 17.32 & 61.18 \\
\midrule
2 & 100 & 44.53 & 25.79 & 12.27 & 27.08 & \textbf{67.19} & 37.97 & 12.24 & 110.55 & 64.84 & 38.51 & 14.77 & 122.64 \\
2 & 200 & 64.06 & 5.68 & 17.23 & 37.99 & 73.44 & 8.06 & 12.95 & 76.51 & \textbf{75.00} & 5.96 & 18.17 & 88.73 \\
2 & 500 & 75.78 & 6.84 & 16.15 & 49.79 & 80.47 & 9.69 & 14.06 & 57.14 & \textbf{82.03} & 4.96 & 14.46 & 65.91 \\
2 & 1000 & 78.13 & 6.43 & 16.87 & 52.21 & 79.69 & 11.50 & 15.85 & 53.67 & \textbf{81.25} & 4.04 & 15.12 & 61.98 \\
2 & 2000 & 72.66 & 10.91 & 22.94 & 54.52 & 79.69 & 10.08 & 14.79 & 54.41 & \textbf{80.47} & 5.94 & 14.97 & 60.62 \\
\midrule
4 & 100 & \textbf{48.44} & 44.56 & 33.95 & 26.02 & 17.19 & 25.23 & 31.09 & 206.01 & 19.53 & 28.30 & 23.09 & 206.45 \\
4 & 200 & 63.28 & 25.31 & 15.01 & 34.52 & 67.97 & 19.21 & 17.75 & 103.86 & \textbf{72.66} & 22.09 & 12.85 & 109.77 \\
4 & 500 & 71.88 & 6.91 & 20.19 & 52.28 & 76.56 & 10.89 & 17.49 & 60.30 & \textbf{79.69} & 9.86 & 15.90 & 69.25 \\
4 & 1000 & \textbf{79.69} & 10.22 & 14.45 & 52.49 & \textbf{79.69} & 7.53 & 16.26 & 55.61 & \textbf{79.69} & 6.29 & 17.09 & 62.16 \\
4 & 2000 & 78.13 & 5.62 & 17.96 & 54.41 & 75.78 & 10.38 & 17.55 & 54.13 & \textbf{78.91} & 7.69 & 16.13 & 62.65 \\
\bottomrule
\end{tabular}%
}
\end{table}

\end{document}